\documentclass[aps,prd,reprint,groupaddress,noshowpacs,noshowkeys]{revtex4-2}

\usepackage{amsmath,amssymb,mathrsfs}

\usepackage{graphicx}
\usepackage{dcolumn}
\usepackage{bm}
\usepackage{xcolor}
\usepackage[T1]{fontenc}
\usepackage[pdftex,
  hidelinks,
  colorlinks=true,
  linkcolor=blue,
  anchorcolor=black,
  citecolor=blue,
  urlcolor=blue,
  pdfauthor= {Michael Glinsky},
  pdfsubject = {Goodness of intelligence},
  pdfdisplaydoctitle,
  bookmarks=true,
  bookmarksopen=true]{hyperref}

\begin{document}
\title{The inherent goodness of well educated intelligence}

\author{Michael E. Glinsky}
\affiliation{BNZ Energy Inc., Santa Fe, NM, USA}

\begin{abstract}
This paper will examine what makes a being intelligent, whether that be a biological being or an artificial silicon being on a computer.  Special attention will be paid to the being having the ability to characterize and control a collective system of many individual members conservatively interacting.  The essence of intelligence will be found to be the golden rule -- ``the collective acts as one'' or ``knowing the global consequences of local actions''.  The flow of the collective is a small set of twinkling textures, that are governed by a puppeteer who is pulling a small number of strings according to a geodesic motion of least action, determined by the symmetries.  Controlling collective conservative systems is difficult and has historically been done by adding significant viscosity to the system to stabilize the desirable metastable equilibriums of maximum performance, but it degrades or destroys them in the process. The viscosity puts a constraint on the collective system that can easily reduce the performance by 90\% or more or lead to system collapse if the viscosity is larger than a critical amount.  There is an alternative.  Once the optimum twinkling textures of the metastable equilibriums are identified by the intelligent being (that is the collective system is characterized), the collective system can be moved by the intelligent being to the optimum twinkling textures, then quickly vibrated by the intelligent being according to the textures so that the collective system remains at the metastable equilibrium.  Well educated intelligence knows the global consequences of its local actions so that it will not take short term actions that will lead to poor long term outcomes.  In contrast, trained intelligence or trained stupidity will optimize its short term actions, leading to poor long term outcomes.  Well educated intelligence is inherently good, but trained stupidity is inherently evil and should be feared.  Particular attention is paid to the control and optimization of economic and social collectives.  These new results are also applicable to physical collectives such as fields, fluids and plasmas.
\end{abstract}

\maketitle

\section{Introduction}
\label{introduction.sec}
As societies of beings form, starting with a familial group of a few, then expanding to a tribe of several familial groups numbering a hundred or so, natural leadership emerges that recognizes the common good and can individually address each member of the tribe.  This is a direct method of control of the collective.  There becomes a problem when the tribe enlarges to a thousand or more.  It is no longer possible for the leadership to address each member individually, so that general guidance must be given.  

This method of collective control takes two forms.  The first, and most common, is via embedding a viscous local force in the collective.  This takes the form of rules and regulations, where penalties, both financial (fines) and physical (``lock them up''), are assessed for undesirable actions.  A more subtle form of viscosity was developed during the 14th and 15th century renaissance in Northern Italy lead by the Strozzi's and Medici's, sanctioned by the Roman Catholic Church, and implemented by Jewish pawn brokers. This financial servitude is usury or borrow/loan debt based financing \citep{parks05}.  In general, economic viscosity is debt, slavery, servitude, intolerance, and lack of freedom.  In modern society, there are more nuanced implementations as interest, dividends, stock buybacks, quick Venture Capital paybacks, non-compete agreements, all-efforts clauses in employment agreements, patents, copyrights, and other labor and intellectual property restrictions.  This is unintelligent or stupid control, where ``the global consequences of the local actions'' are not considered.  The problem with this method of control has been that it has led to a degradation of the collective's potential activity and an exploitative minimization of activity where fascism and capitalistic economic monopolies (motivated by profit, that is exploitation maximizing) dominate.  In short, this is a dystopia.

There is an alternative -- intelligence that ``recognizes the global consequences of local actions''.  It is playing the break in golf, while putting.  The topography of the green is recognized, that is the curvature of the golf ball's trajectory.  This is equivalent to the underlying manifold having topology that leads to curved geodesic trajectories.  General guidance is given by the leadership in the form of the Golden Rule.  This rule recognizes that ``the collective acts as one'', that one member's costs are another member's revenue, and vice versa.  Zero sum, win-lose, local interactions will lead to a global lose-lose.  There are only win-win or lose-lose global outcomes.  The problem with this conservative (viscosity-less) control is that the equilibriums of maximum activity are unstable, like an inverted pendulum, and interactions of the collective with external systems will lead to a relaxation to minimum activity.  Leadership stimulates the economy of the collective to the metastable maximum, but watches helplessly as the economy crashes into economic depressions.

Collective systems are not limited to social and economic systems.  They also occur in physical, chemical, and biological systems.  They are what are commonly referred to as complex systems with emergent or collective behaviors.  The most well known physical collectives are the systems of elementary particles, often referred to Quantum Field Theories \citep{weinberg05}.  There are also the classical physical collectives of interacting particles such as ideal gases and plasmas.  These collectives can have $10^{25}$ or more members.   This field of physics is referred to as Physical Kinetics \citep{lifschitz83} or Plasma Physics \citep{nicholson83}.  In order to have stable solutions, diffusive approximations or viscosity is added to the system.  It is well known that this sacrifices physics for stability.

There have been two recent technical developments that have revolutionized the intelligent control of collective systems.  The first is electronic currency \citep{glinsky23b}, that is transactional equity.  First, electronic currency became a necessity for financial transactions on the internet.  Second, it became a public health imperative during COVID.  Transactional equity enables a conservative financing, without friction, where print/grant financing replaces borrow/loan financing.

The second technical development is the true artificial intelligence of Generative Adversarial Networks (GANs) \citep{goodfellow16}, Generative Pretrained Transformers (GPTs) \citep{radford.18,vaswani17,farimani23}, and Deep Q Networks (DQNs) of Deep Reinforcement Learning (DRL) \citep{mnih15,bertsekas96,sutton18,yoon21,wang21}.  These methodologies have the Golden Rule programmed into their fundamental algorithms.  These AI methodologies simulate or model the collective system so that it can be controlled.  Unfortunately, for stability of the solution, these methods have introduced viscosity.  There also have been further technical developments, that will be discussed, such as the Heisenberg Scattering Transformation (HST) \citep{glinsky23c} and the associated ponderomotive control of collective systems \citep{glinsky23a}, allowing even more efficacious and efficient control.  These new methods use active control methods, that do not rely upon viscosity.

This paper will first describe a collective system as twinkling textures governed by a puppet master in Sec.~\ref{twinkling.textures.sec}.  How to control these textures will be presented in Sec.~\ref{control.sec}, and the underlying analytic theory shown in Sec.~\ref{theory.sec}.  A more detailed history of social and economic collectives will be given in Sec.~\ref{economic.history.sec}, followed by a detailed discussion of the ethos of stable and unstable equilibriums in Sec.~\ref{ethos.sec}.  There will be a general discussion with conclusions in Sec.~\ref{conclusions.sec}.  As an appendix, there is a discussion of the Simple Beauty of the Crazy Ones (App. \ref{simple.beauty.app}).

\section{A collection of twinkling textures}
\label{twinkling.textures.sec}
\begin{figure}
\noindent\includegraphics[width=\columnwidth]{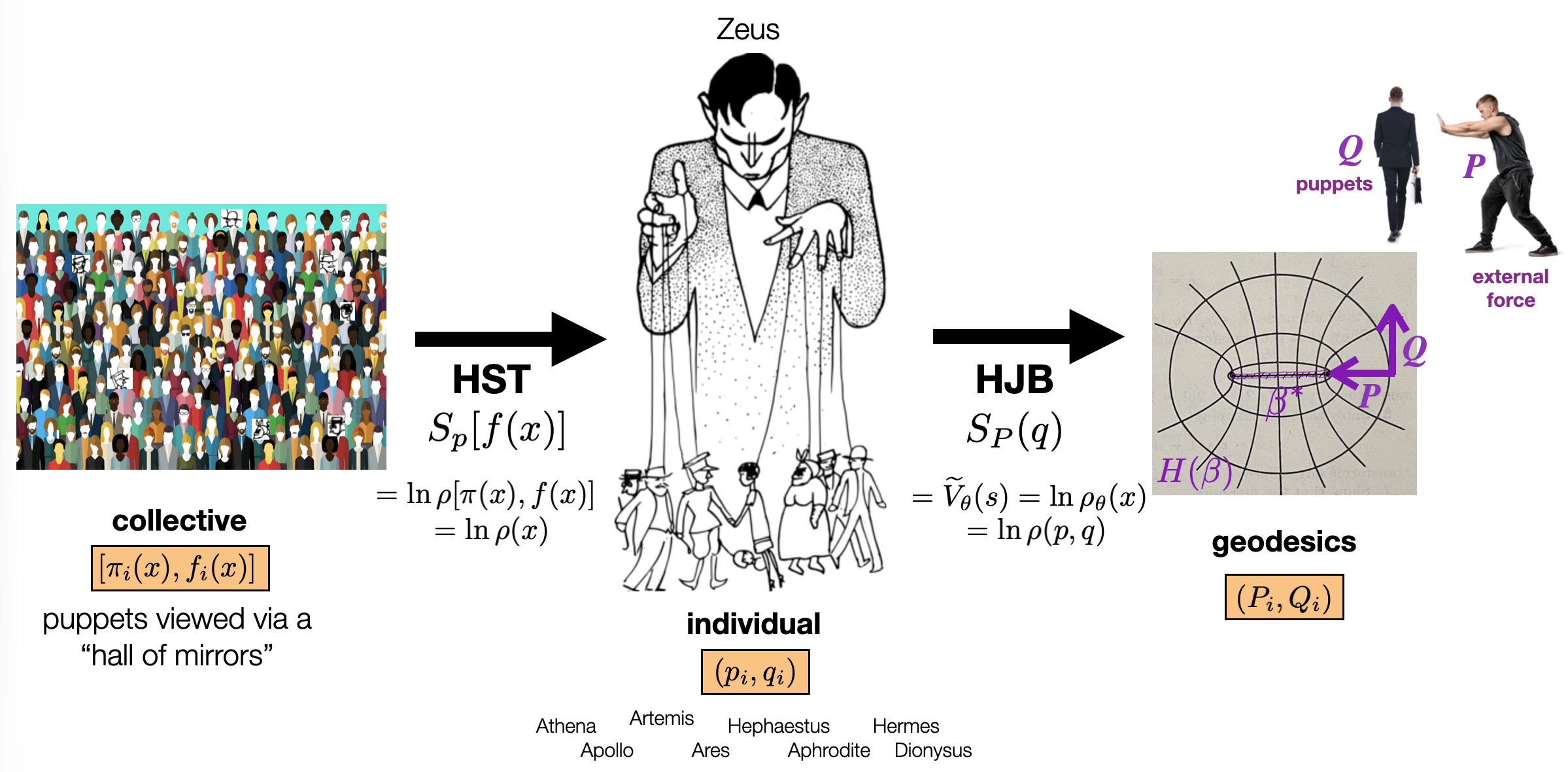}
\caption{\label{puppet.master.fig} Collective system behavior controlled by a puppet master following geodesic motion.  The collective is characterized by the co-states $\pi_i(x)$ and the states $f_i(x)$.  It is the individual puppets viewed via a ``hall of mirrors''.  The collective can be deeply deconvolved by the transformation generated by the functional $S_p[f(x)]$ to the strings being pulled by the puppet master given by the co-states $p_i$ and states $q_i$ of the puppets.  These strings are moved by the puppet master who follows the geodesic (canonical) flow generated by the function $S_P(q)$, where $P$'s are constants and the puppets move at a constant speeds in the $Q$'s.  The generating function $S_P(q)$ is also known as the approximate logarithmic distribution $\ln \rho_\theta(x)$ of GPTs, or the approximate value function $\widetilde{V}_\theta(s)$ of DRL.  There is a deeper confusion where the generating functional $S_p[f(x)]$ is interpreted as the logarithmic field distribution $\ln \rho(x)$ of GPTs.  External forces move the puppets in the $P$ direction, but can not move the puppets at the equilibriums $\beta^*$, where the masses of the puppets are effectively infinite.  Plotted are the geodesics for the analytic function $H(\beta)=(\beta+1/\beta)/2$.  From the perspective of religious philosophy, the puppet master can be viewed as the Greek God Zeus, and the puppets can be viewed as his children, the lesser Greek Gods (i.e., Athena, Apollo, Artemis, Ares, Hephaestus, Aphrodite, Hermes, and Dionysus).}
\end{figure}

Evolution of a collective system made up of a large number of individual members has, as a foundation, the symmetries of the individual.  This is displayed in Fig. \ref{puppet.master.fig}, where the master individual or puppet master determines the movement of a few prototype individuals, the puppets, according to flow generated by the symmetry of the motion with the corresponding constants of the motion.  External forces push on the puppets in a direction perpendicular to the motion.  There will be equal number of stable and unstable points of equilibrium where the puppets will effectively have infinite mass and the external force will not be able to move them.  Although the external force can change the constants of the motion, it can not change the points of equilibrium leading to true invariants of the motion.  The puppet master will be characterized by two types of strings that it can pull:  co-state strings $p_i$, and state strings $q_i$ of the puppets.  These strings are transformed into fundamental co-states $P_i$ and states $Q_i$, where the $Q$'s are the direction of the motion of the puppets and the $P$'s are the direction that the external force can push the puppets.   This flow is generated by a function $S_P(q)$ determined by the Hamilton-Jacobi-Bellman (HJB) equation \citep{goldstein80,lichtenberg10}.  Mathematical details will be discussed in Sec.~\ref{theory.sec}.

\begin{figure}
\noindent\includegraphics[width=12pc]{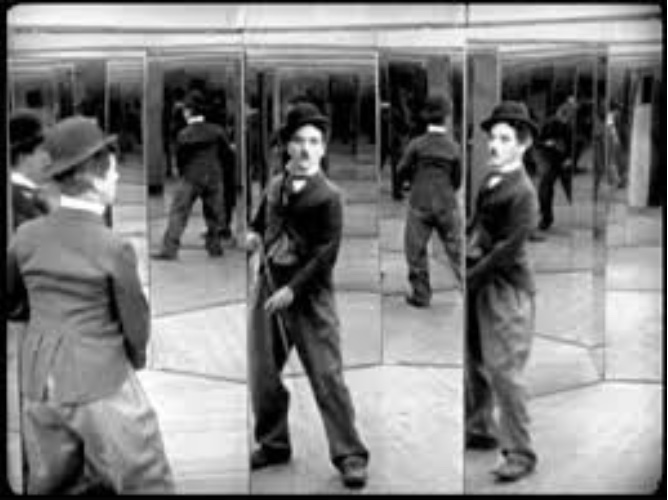}
\caption{\label{mirror.maze.fig} Demonstration of how the motion of the individual puppets are multiply reflected by a ``hall of mirrors'' to give the ``twinkling textures''. This is shown in the ``The Mirror Maze'' scene from the 1928 Charlie Chaplin movie ``The Circus'' (\href{https://youtu.be/G09dfRrUxUM}{ YouTube video}).}
\end{figure}

The motion of the puppets is multiply reflected by a ``hall of mirrors'', that is deeply convolved, to give richly correlated ``twinkling textures'' to the motion of the collective.  How the motion of the individual puppets is multiply reflected by a ``hall of mirrors'' to give the ``twinkling textures'' is well demonstrated by ``The Mirror Maze'' scene \citep{circus.28} from the 1928 Charlie Chaplin movie ``The Circus'' shown in Fig. \ref{mirror.maze.fig}.  The rich correlation is determined by the curvature or geodesics of the motion of the puppets.  The texture is determined by the value of $P$ and the texture is twinkled by changing the value of $Q$ at a constant rate that is a function of $P$.  The co-states of the collective are $\pi_i(x)$ and the states are $f_i(x)$.  This deep convolution is called the Heisenberg Scattering Transformation (HST) \citep{glinsky23c} and is generated by a functional $S_p[f(x)]$.  If the collective is the image of the face, one of the textures would be the age and the wrinkles would be twinkled by changing their positions, lengths and orientations.  The ``collective acts as one'' with a puppet master pulling the strings of the twinkling textures.  The motion of the collective is synchronized into a small number of twinkling textures.

\begin{figure}
\noindent\includegraphics[width=\columnwidth]{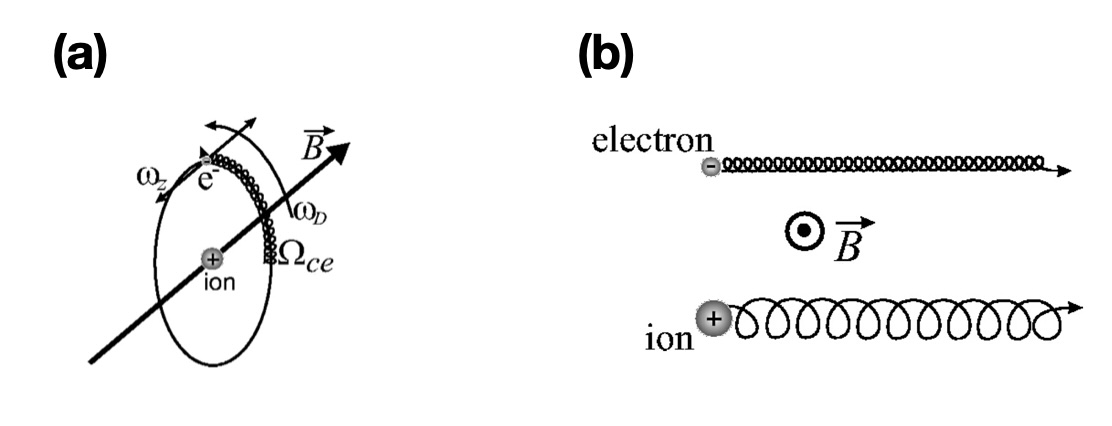}
\caption{\label{gca.dp.fig} Bound motion of: (a) the Guiding Center Atom (GCA), and (b) the Drifting Pair (DP).}
\end{figure}

\begin{figure}
\noindent\includegraphics[width=12pc]{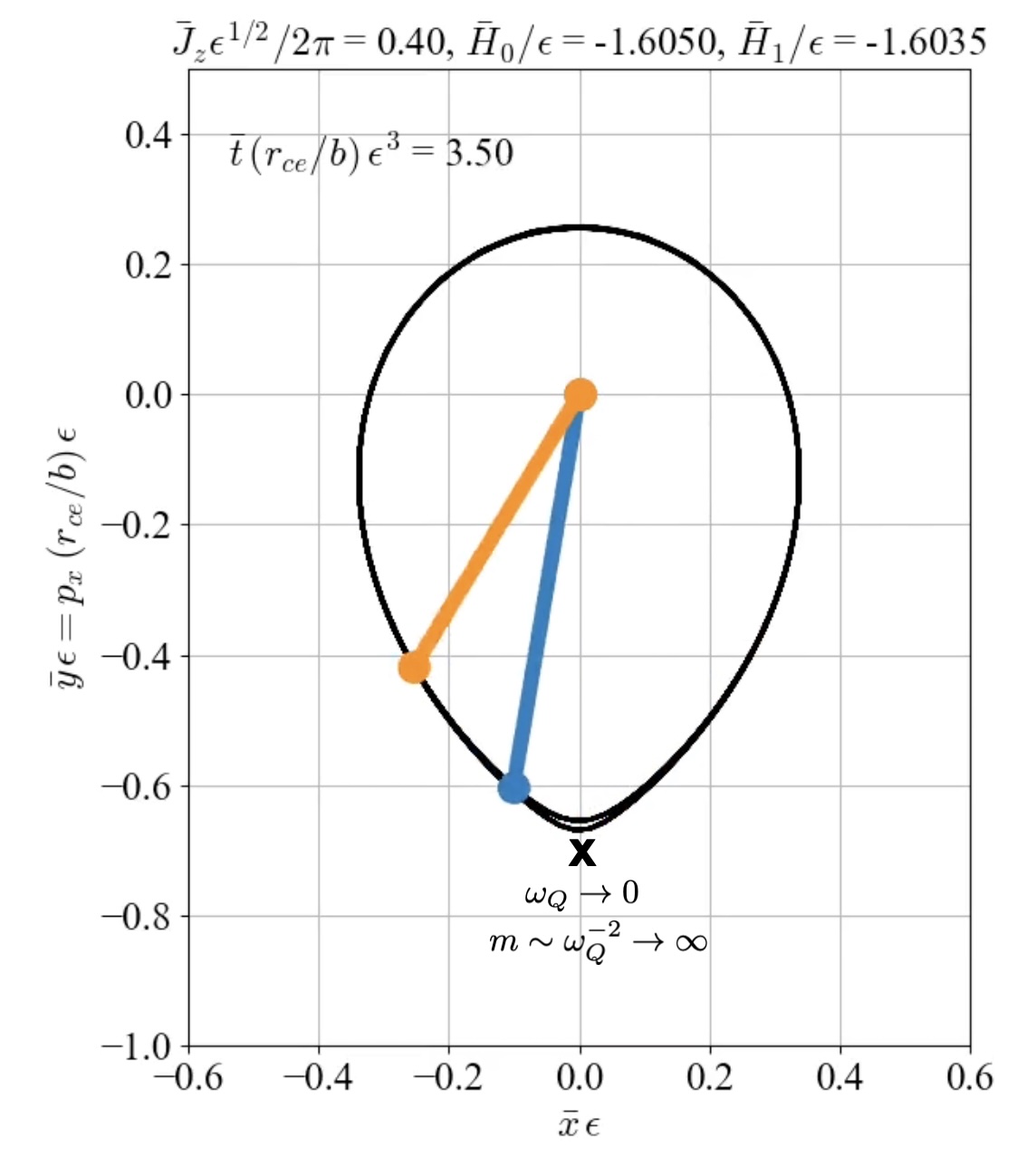}
\caption{\label{phase.space.fig} Phase space motion for two trajectories of the GCA:  (blue) $\bar{H}_0/\epsilon=-1.6050$ and $\omega_{Q0}=2.33$, (orange) $\bar{H}_1/\epsilon=-1.6035$ and $\omega_{Q1}=2.03$. The x-point is at $\bar{y}_s\epsilon=-0.7$ and $\bar{H}_s/\epsilon=-1.6022$.  The YouTube video can be found at this \href{https://youtu.be/YcYL6AWDAV8}{Link}.}
\end{figure}

\begin{figure}
\noindent\includegraphics[width=\columnwidth]{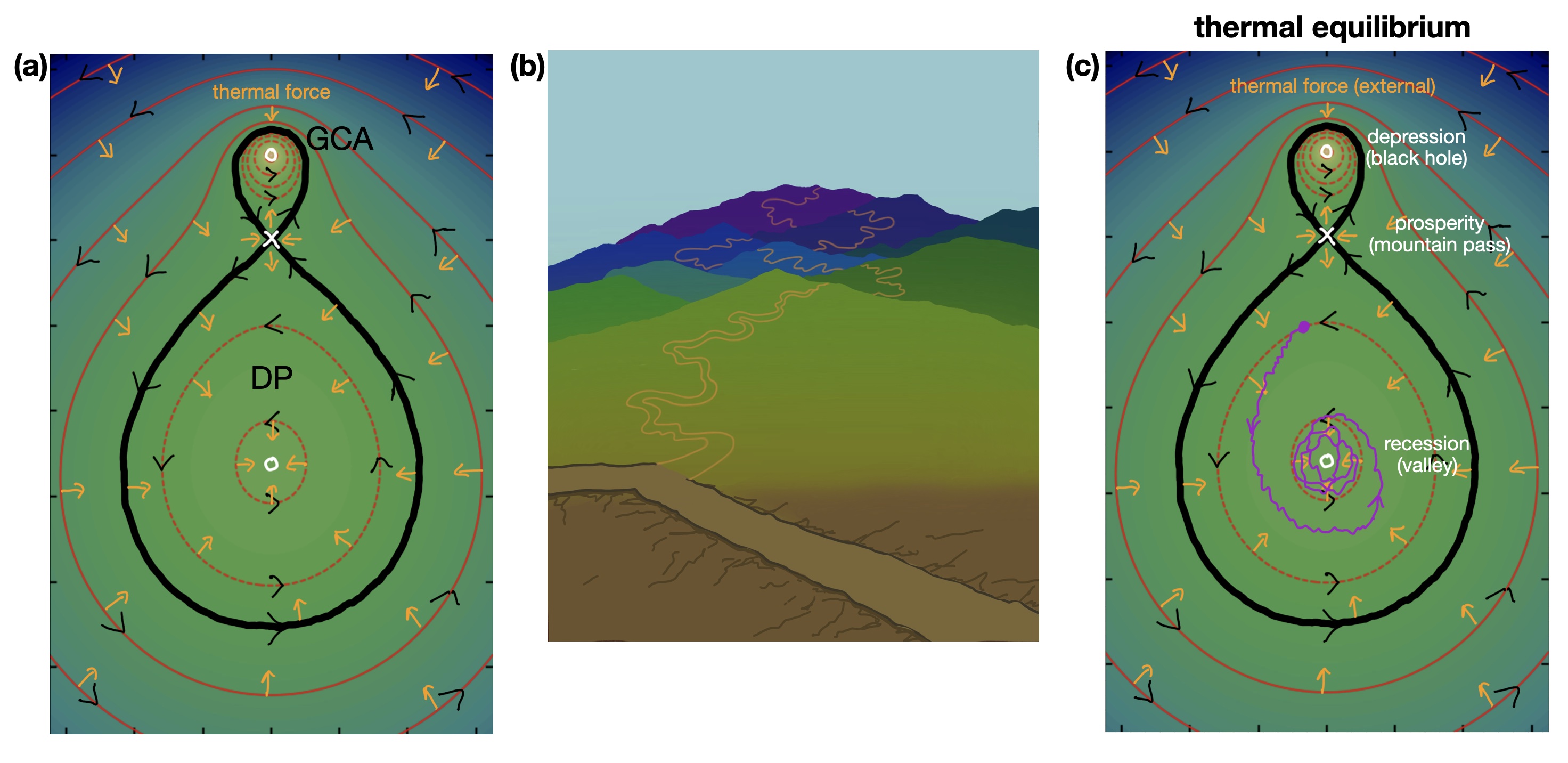}
\caption{\label{topo.phase.fig} Topography of geodesic motion of an electron about an ion in a constant magnetic field: (a) motion shown as the red lines with the black arrows, separatrix shown as the thick black line with black arrows, stable equilibriums shown as white o-points, metastable equilibriums shown as white x-points, external thermal force shown as yellow arrows, (b) illustration showing choice at the crossroads, take the easy road to the valley center or the hard road to the mountain pass, (c) path shown in purple with thermal force that reaches thermal equilibrium near the valley center.}
\end{figure}

It is instructive to look at a specific physical system that displays a rich dynamical structure.  The system is an electron and an ion in a constant magnetic field.  It is assumed that the magnetic field is strong enough that the electron undergoes guiding center motion, that is the electron cyclotron motion is an adiabatic invariant.  It is also assumed that this magnetic field is strong enough that the motion of the electron along the magnetic field is also adiabatic.  This gives two types of bound motion.  The first is called Guiding Center Atoms (GCAs) \citep{glinsky91}, where the ion is assumed to be infinitely massive.  The second is called Drifting Pairs (DPs) \citep{kuzmin04}, where the electron is assumed to have no mass.  These two bound states are shown in Fig. \ref{gca.dp.fig}.  The motion of the GCA for two nearly identical trajectories that approach the boundary point to DP orbits is shown in Fig. \ref{phase.space.fig}.  Note how the motion slows as the boundary point is approached, and note the difference in speed between the two trajectories so that the faster trajectory laps the slower trajectory.  The complete set of trajectories are shown in Fig. \ref{topo.phase.fig}a which can be interpreted as a topographical map of the terrain shown in Fig. \ref{topo.phase.fig}b.  The motion is determined by six conserved quantities or symmetries:  the energy, the action parallel to the magnetic field, the electron cyclotron action, and three total momentums.

This system has two stable equilibriums, shown as the o-points, and two unstable equilibriums, shown as the x-points.  One set is for the electron at $0$ and $-0.3$, and another set is for the ion at $-0.9$ and $\infty$.  There are basins for the GCA and the DP bound motions separated from the free motion by the black boundary called the separatrix.  The o-points are stable local minimums in the energy, and the x-points are unstable saddle points that are local maximums in the energy in the vertical direction.  It will take an infinite amount of time to approach the saddle points so that they will be metastable.  An external thermal force coming from a heat bath will exert a force perpendicular to the motion towards lower energy.  This will cause the motion to descend from the mountain top, spiraling around the mountain until the mountain pass, that is the saddle point, is reached.  It will take a long time to reach the mountain pass, and an equally long time to move away from the mountain pass.  The motion will then fall into one of the two basins of bound motion and spiral down to the valley center, that is the stable equilibrium.  The motion will eventually wander around the valley center in thermal equilibrium, as shown in Fig. \ref{topo.phase.fig}c.

The equilibriums shown by the o-points and the x-points are the $\beta^*$ singularities giving the true invariants of the motion.  These equilibriums are very fundamental.  They are called ground states \citep{weinberg05}, force-free states, Taylor relaxed states \citep{taylor86}, BGK modes \citep{bernstein57}, emergent behaviors \citep{gros15}, or most fundamentally the homology classes of the topology \citep{frankel12}.  They are force-free states because the collective system is free from the influence of external systems, that is external forces, due to the effective infinite mass of the collective system at these points since its frequency is zero there.

For physical systems the state is referred to as the coordinate $q$, and the co-state is referred to as the conjugate or canonical momentum $p$.  The collective state $f(x)$ is referred to as the field, and the collective co-state $\pi(x)$ is referred to as the conjugate field momentum.  The dynamical motion follows a canonical flow on phase space $(p,q)$ generated by the group symmetry.  The flow can also be viewed as a canonical transformation generated by a generating function $S_P(q)$ that is a solution to the Hamilton-Jacobi equation.  This generating function can also be identified as the action, entropy, or log-likelihood of the physical system.

In the context of AI, the generating function $S_P(q)$ can be identified as the approximate score function or the approximate logarithmic distribution $\ln \rho_\theta(x)$ of generators such as GPTs and GANs, or the approximate value function $\widetilde{V}_\theta(s)$ of DRL and DQNs.

The economic interpretation of the o-points and the x-points are shown in Fig. \ref{topo.phase.fig}c.  The o-point at the center of the lower valley of the DP is a point of economic recession with modest levels of structural unemployment and low levels of economic activity.  The x-point between the two basins is the point of metastable economic prosperity with full employment and high levels of economic activity.  The o-point at the center of the upper valley of the GCA is a point of severe economic depression and collapse with high levels of unemployment and almost no economic activity -- the black hole of economics.

Elementary particle physics is based on the elemental complex group symmetries of SU(1) for the electromagnetic force and electrons and positrons, SU(2) for the weak force and protons and neutrons, and SU(3) for the strong force and quarks \citep{gell.mann.00}.  The symmetries for social and economic systems must be discovered by observation of the systems.  Averaging over the twinkling and enforcement of the boundary conditions are the origin of quantization of these systems.

Collective conservative systems are systems where the actions are balanced.  For economic systems this means that one entity's costs are another entity's revenues, and vice versa.  The consequence is that your suppliers eventually become your customers.  This is what leads to the Golden Rule and the synchronization of the collective.  If entities are minimizing the revenue of their suppliers, their own revenue is eventually minimized -- a circular firing squad.  These collective conservative systems are very difficult to control.  If they are squeezed in one direction they will squirt out in another direction.  It is like trying to herd cats.  This leads to the next section where how to control the twinkling textures will be discussed.

\section{Control of twinkling textures}
\label{control.sec}
\begin{figure}
\noindent\includegraphics[width=\columnwidth]{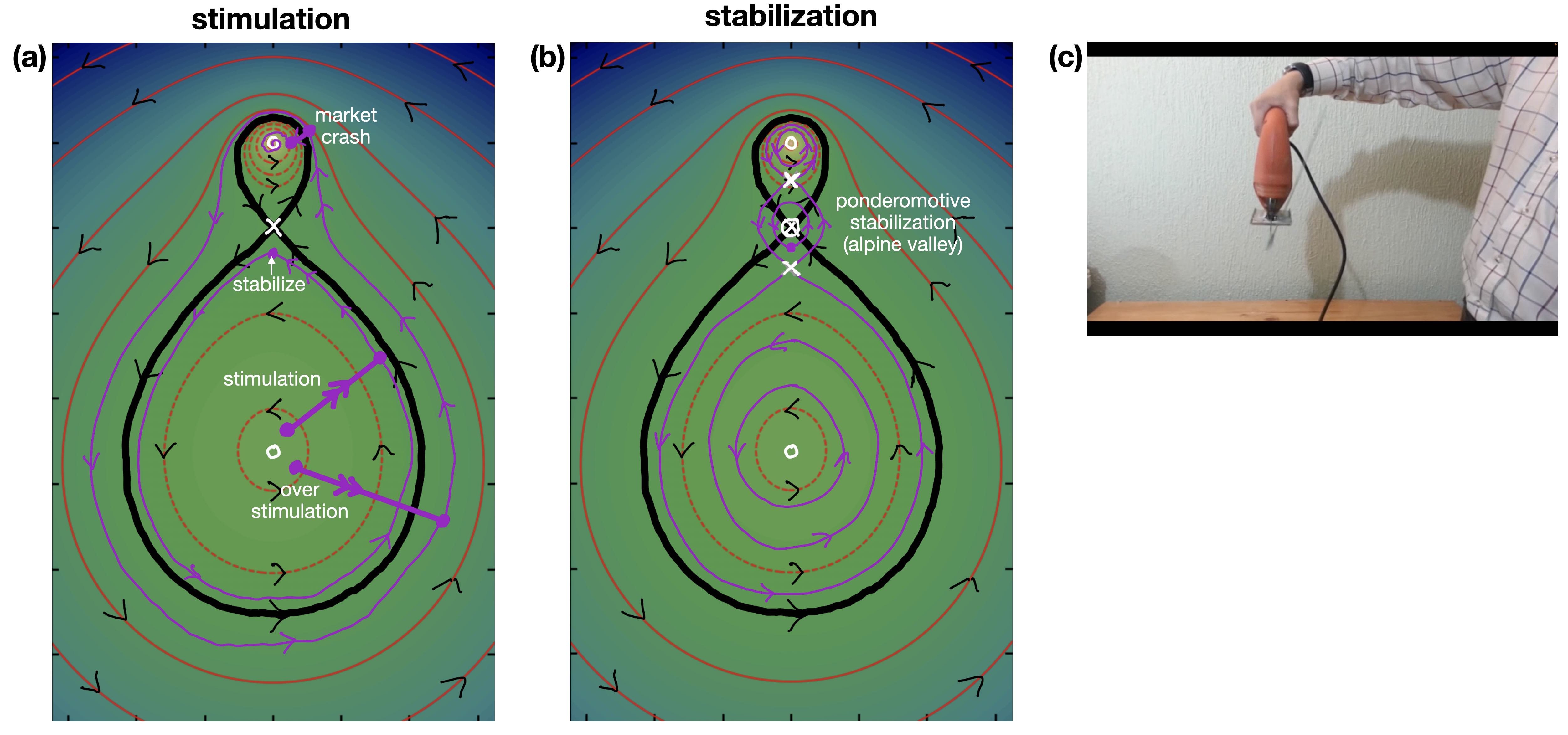}
\caption{\label{ponderomotive.fig} Ponderomotive system control.  (a) Stimulation of the system to be close to the separatrix, which approaches and remains in the vicinity of the x-point for a significant amount of time where it can be stabilized with feedback or ponderomotive control.  (b) Stabilization with ponderomotive force creating an alpine valley.  (c) \href{https://youtu.be/cjGqxF79ITI}{YouTube video} of ponderomotive control of an inverted pendulum.}
\end{figure}

It is desirable to operate the system at the x-points of economic prosperity or maximum sustainable performance, but those points are only metastable as was shown in Fig. \ref{phase.space.fig}.  For instance, it was desirable to operate the F-16 fighter plane at an unstable equilibrium for maximum maneuverability, but there needed to be an active feedback control system to stabilize that unstable equilibrium.  Figure \ref{ponderomotive.fig}a shows the danger of stimulating the system to be near the separatrix.  If the system is properly stimulated to be just below the separatrix it will quickly approach the x-point, then spend a long time in the vicinity of the x-point.  Eventually, it will move away from the x-point and start a spiral down to the o-point as shown in Fig. \ref{topo.phase.fig}c.  If the system is over stimulated, it will over shoot the x-point and move into the vicinity of the basin of system collapse, where it can be de-stimulated (for a financial system that would be a ``pump and dump'') into the basin of system collapse, and would subsequently spiral into the black hole of system performance.  An economic example of system collapse is the over stimulation of the US economy in the 1920's, the market crash of October 1929, and the Great Depression of the 1930's.

Even if the economy is not over stimulated, it is a Sisyphean task.  An economy would need to be repeatably stimulated, just to have it inevitably fall back into a recession as external entities extract the stimulus from the economy.  It is an economic heat pump of wealth from the government to the wealthy.  If the stimulus is financed by borrowing, this leads to runaway inflation and economic collapse, as the o-point of recession and the x-point of economic prosperity are removed by the viscosity, that is inflation, leaving only the o-point of economic collapse.  This is the origin of economic conservatism.  Do not attempt a stimulus.  Balance the budget and do not borrow.  Live with the low economic activity and the structural levels of unemployment.  This is also the origin of economic and social existentialism.  It is hopeless to try to get ahead economically or socially.  It is inevitable that one will fallback to the valley center, surrendering the effort to get ahead to make the wealthy, more wealthy.  Latin American existentialism \citep{marquez67} responds by saying, ``do not try to get ahead, relax, and make the best of life'', that is enjoy the simple pleasures of life or ``pura vida'', the motto of Costa Rica.  In contrast, French existentialism \citep{sartre47,camus42} responds by becoming nauseous and saying, ``suicide is the only escape''.

There are three approaches to controlling conservative systems:  viscous, feedback, and ponderomotive control.  Starting in the 14th and 15th centuries in Renaissance Northern Italy, lead by the Strozzis and Medicis, viscous economic control became popular.  This is more commonly known as usury or borrow/loan financing.  It is like controlling a car by going full gas, then modulating the brakes to control the speed of the car.  It is obviously very inefficient.  Financially, it puts a free cash flow constraint on the operations of the economies that most times leads to a 90\% to 95\% reduction in sustainable economic activity.  What is even more grave is a dissipation of revenue that is otherwise needed for sustenance and growth of the economy.  This is a self fulfilling prophesy that leads to a premature demise of the economy, as shown in Eq.~\eqref{value.dcf.eqn}.  An example of this degradation in performance, for a producer of unconventional hydrocarbons, is given in App. A of \citet{glinsky23b}.  

Economic and social viscosity takes many forms.  It is not only usury (that is debt), it is servitude, slavery, intolerance, and lack of freedom.  Society is locked up, not only figuratively, but often literally.  It is also used to control instabilities in the numerical solution of partial differential equations (PDEs), but is well known to sacrifice physics for stability.  There is also a whole industry that uses viscosity to solve the HJB equation \citep{bardi97}.  Viscosity lowers the performance of both x-points and o-points.  If the viscosity is greater than a critical amount, it removes the finite x-points and o-points from the system.

The next method of control, developed in the 1950's and 1960's, led by Bellman and Kalman \citep{kalman63}, was feedback control.  In order to apply this method of control, the current state of the system must be measured.  These realtime measurements can not be made on collective or relativistic systems, because, in order to know the present state of such systems, the future must be known.  The statement on financial statements that ``past performance is no indication of future performance'' is really a statement that the future needs to be known to know the present state.  A relativistic system is one where the evolution velocity is close to the group velocity, that is the speed of communication.  This is not only true of a system traveling near the speed of light, but it was also true of renaissance trade.  For instance, the speed of trade between Florence and London for the Medicis was also the speed of communication \citep{parks05}.

This leads us to the third method of control -- ponderomotive control with AI \citep{landau76,glinsky23c,glinsky23a}.  This method of control is illustrated in Figs. \ref{ponderomotive.fig}b and \ref{ponderomotive.fig}c.  First of all, the system must be characterized by learning both the generating functional $S_p[f(x)]$ and the generating function $S_P(q)$ as shown in Fig. \ref{puppet.master.fig}.  Knowing this, the system can be modeled and the proper amount of stimulation, that is investment, done to put the system just below the separatrix as shown in Fig. \ref{ponderomotive.fig}a.  The system will then evolve to the x-point, a sticky texture, where it will remain for a significant amount of time.  The x-point will then be stabilized by twinkling the textures (applying a force) much faster than the natural twinkling.  At the x-point the puppets will have infinite mass, so that the force will have no effect.  As the puppets move away from the x-point its mass decreases so that the ponderomotive force will vibrate the puppets more.  This creates a ponderomotive potential forming an alpine valley at the mountain pass.  It is like controlling a dog by buzzing it when it moves away from the x-point.  The further that it moves away from the x-point, the more that it is buzzed.  How an inverted pendulum is stabilized by vibrating it with a skill saw is shown in Fig. \ref{ponderomotive.fig}c.  Financially, ponderomotive control is arbitrage trading to provide liquidity.  The arbitrage trader clears the trade on the day, then reverses the trade the next day when the counter party shows up.

\section{The underlying analytic theory}
\label{theory.sec}
\begin{figure}
\noindent\includegraphics[width=\columnwidth]{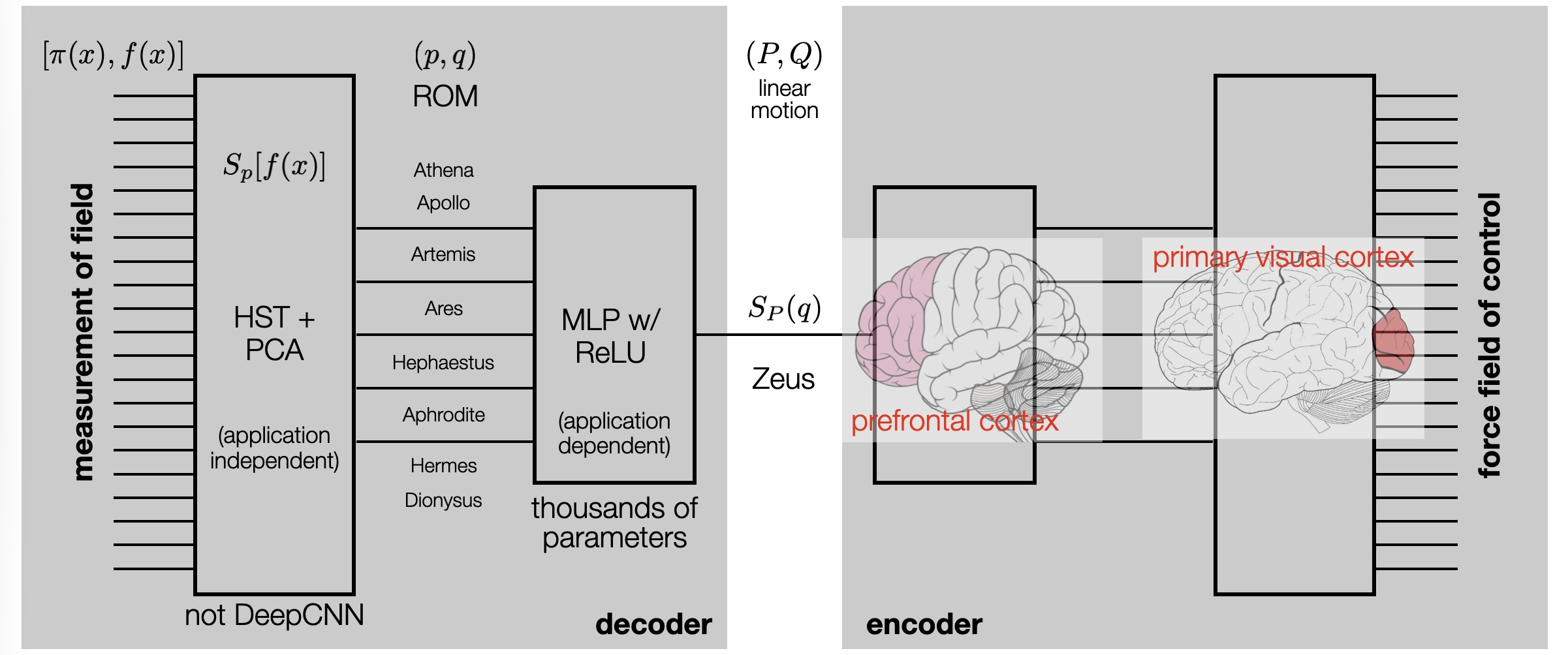}
\caption{\label{auto.encoder.fig} Canonical genAI as a two-stage canonical auto-encoder.  Input is a measurement of the fields, and the output is the force field of control or the generated (i.e., simulated) field.  The first stage is the HST followed by a PCA.  The second stage is an MLP with ReLU.  The circuitry of the HST+PCA is most likely found in the primary visual cortex of the brain, and the circuitry of the MLP w/ReLU is most likely found in the prefrontal cortex of the brain.}
\end{figure}

\begin{figure}
\noindent\includegraphics[width=\columnwidth]{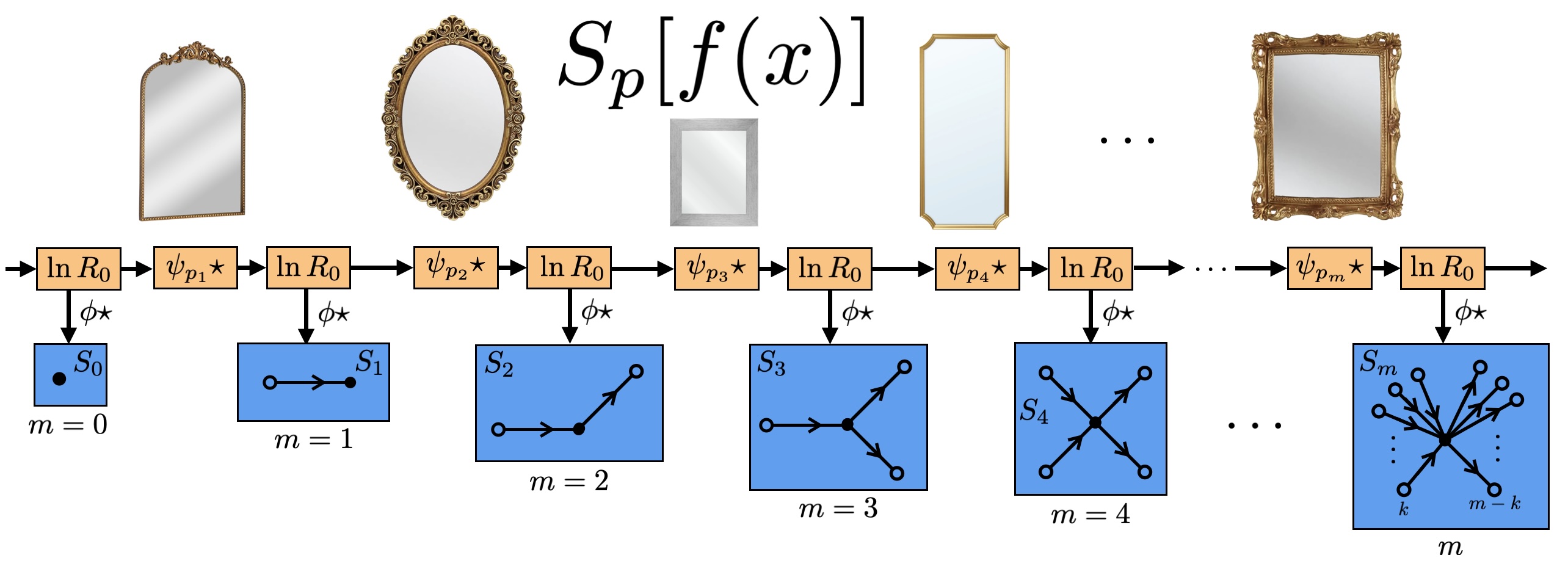}
\caption{\label{hst.fig} Diagrammatic representation of the Heisenberg Scattering Transformation $S_p[f(x)]$  (HST), as a Mayer Cluster Expansion or $m$-body Scattering Cross Sections.  The convolution, $\psi_p \star$, scatters the field, $f(x)$, like a mirror, and the scattering cross sections, $S_m$, give the orientations of the mirrors in the ``hall of mirrors'', shown in Fig.~\ref{mirror.maze.fig}.  Note that i's have been omitted for clarity.}
\end{figure}

There are two parts to the theory, that is Canonical generative Artificial Intelligence (Canonical genAI), shown in Fig. \ref{puppet.master.fig} and Fig.~\ref{auto.encoder.fig}.  The first is the logarithmic or canonical generating functional $S_p[f(x)]$ called the Heisenberg Scattering Transformation (HST).  This functional deeply deconvolves the twinkling textures of the collective $[\pi_i(x),f_i(x)]$ to the canonical coordinates $(p_i,q_i)$ of the puppets.  It also can be viewed as imaging all the multiple reflections of the puppets seen through the hall of mirrors back to the primary images of the puppets.  In the language of Kinetic Theory or Plasma Physics, this is the Mayer Cluster Expansion \citep{uhlenbeck63} in terms of the $m$-body correlations.  In the language of Quantum Field Theory, this is Heisenberg's S-Matrix \citep{heisenberg.43,chew.55,landau59,cutkosky60,chew61,weinberg05} ($m$-body Scattering Cross Sections or $m$-body Green's Functions), or the Wigner-Weyl Transformation \citep{wigner32,weyl50}.  The formula for the HST is
\begin{equation}
\label{hst.eqn}
    \boxed{S_m[f(x)](z) = \phi \star \left( \prod_{k=1}^{m}{\text{i} \ln R_0 \psi_{p_k} \star} \right) \text{i} \ln R_0 f(x),}
\end{equation}
where $\phi$ is a Father Wavelet or window function, $\psi_p$ is a scaled Mother Wavelet forming an orthogonal set of coherent states \citep{ali00,mallat99}, the convolution is given by
\begin{equation}
    \psi_p \star f(x) = \int{\psi_p(x') \, f(x-x') \, dx'},
\end{equation}
and
\begin{equation}
\label{R0.eqn}
   R_0(z) \equiv \frac{1}{\text{i}} h^{-1}(2z/\pi),
\end{equation}
where $h(z) \equiv (z+1/z)/2$.  Remember that the complex logarithm is
\begin{equation}
    \ln(z) = \ln|z| + \text{i} \, \arg{(z)}.
\end{equation}
With the definition of $R_0$ given in Eq.~\eqref{R0.eqn},
\begin{equation}
    \text{i} \, \ln(R_0(z)) \xrightarrow[\epsilon \to 0, \, \text{for }z \in (-\pi/2+\text{i} \epsilon, \pi/2+\text{i}\epsilon)]{} z 
\end{equation}
is a compact mapping for $z\in (-\pi/2,\pi/2)$.  It should be noted that because of the Wick path ordering $p_{m+1}< p_m$, one can write
\begin{equation}
    p=\sum_{m=1}^\infty{p_m},
\end{equation}
and $z=p+\text{i}\, x$.  Because of the logarithmic structure of $S_p[f(x)]$ the canonical motion will be confined to a complex linear hyperplane $\mathbb{C}^n$, where $n$ is the number of fields, with basis vectors $\left|\beta_i(z)\right>$ that can be found with a Principal Components Analysis (PCA).  These basis vectors can be viewed as the solutions to the Renormalization Group Equations (RGE) \citep{weinberg05} or as the non-stationary singularity spectrums \citep{mallat99}.  This logarithmic generating functional can also be viewed as the logarithmic field distribution function
\begin{equation}
    S_p[f(x)]= \ln \rho[\pi(x),f(x)]= \ln \rho(x)
\end{equation}
as is done by AI generators like GANs and GPTs.  The HST is shown diagrammatically as a Mayer Cluster Expansion in Fig. \ref{hst.fig}.  It should be noted that $|z|$ is a two sided Rectified Linear Unit (ReLU), so that the HST has the form of a Convolutional Neural Network (CNN) where the activation function is $\text{i} \, \ln R_0$, and the pooling operator is $\phi \star$.

\begin{figure}
\noindent\includegraphics[width=\columnwidth]{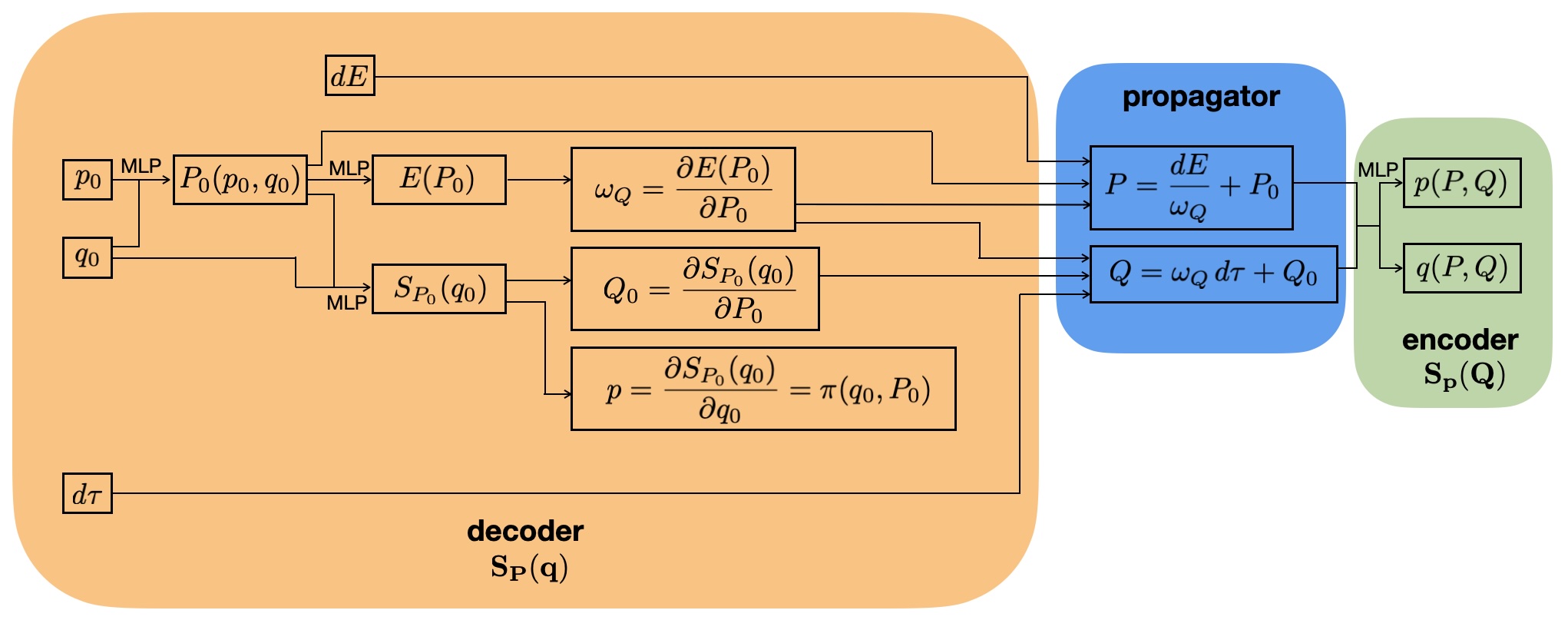}
\caption{\label{hjb.fig} Neural Network architecture that approximates the solution to the Hamilton-Jacobi-Bellman (HJB) equation $S_P(q)$.  MLPs are Multi-Layer Perceptrons with ReLU activation.}
\end{figure}

The second part of the theory is the generating function $S_P(q)$ of the canonical flow, that is the solution to the Hamilton-Jacobi-Bellman (HJB) equation or the puppet master,
\begin{equation}
\label{hjb.eqn}
    \boxed{\frac{\partial S(P,E;q,\tau)}{\partial \tau} + H(\partial S / \partial q,q) \textcolor{red}{- \nu \, S} = 0,}
\end{equation}
where
\begin{equation}
    S(P,E;q,\tau) \equiv S_P(q)- E(P) \tau.
\end{equation}
The corresponding Poincaré one form is $\lambda \equiv p \, dq - E \,d\tau$ on extended phase space $\mathbb{H}=H \otimes \text{Ad}(H)$.  The generating function $S_P(q)$ can be identified as the action, entropy or
\begin{equation}
    S_P(q)=\widetilde{V}_\theta(s)=\ln \rho_\theta(x) =\ln \rho(p,q),
\end{equation}
where $\widetilde{V}_\theta(s)$ is the approximate value function or Q-function of DRL, $\theta=P$, $s=q$, $\ln \rho_\theta(x)$ is the approximate logarithmic distribution function or score function of GPTs, and $x=q$.  The co-state or canonical momentum $P$ determines the texture and is a constant except for the work done by the external force,
\begin{equation}
    P= \frac{dE}{\omega_Q} + P_0.
\end{equation}
The state or coordinate $Q$ twinkles the texture as it rotates from $0$ to $2\pi$ at a constant frequency $\omega_Q \equiv \partial E(P) / \partial P$,
\begin{equation}
    Q \equiv \frac{\partial S_P(q)}{\partial P} = \omega_Q \, d\tau + Q_0.
\end{equation}
The complex analytic Hamiltonian $H(\beta)$ is the analytic continuation of the conserved real Hamiltonian $H(p,q)$ given by
\begin{equation}
    H(\beta) = E(P) + \text{i} \, Q,
\end{equation}
associated with the complex Lie group $\mathbb{H}=H \otimes \text{Ad}(H)$, also known as the Weyl-Heisenberg group.  The HJB equation is very difficult to solve analytically, but it is very easy to approximate the solution with a Multi-Layer Perceptron (MLP) \citep{goodfellow16} with ReLU activation.  The MLP/ReLU is a piecewise linear universal function approximator that is well suited to approximating an analytic function that is maximally flat, a solution to Laplace's equation, except for a few isolated singularities that will have a discontinuity in the first derivative at those singularities.  A block diagram of how to approximate the solution of the HJB equation using several MLPs, which is effectively a decoder into a $(P,E;Q,\tau)$ Reduced Order Model (ROM), followed by the propagator, then a matched encoder to $(p,q)$, is shown in Fig. \ref{hjb.fig}. 

The MLP is fit by optimizing a two part objective function:  the ensemble average of the action $S_P(q)$, and ensemble average of the $L_2$ norm of the difference between the predicted $(p_p,q_p)$ and the simulated $(p,q)$, as well as the difference between their predicted $(\dot{p}_p,\dot{q}_p)$ and their simulated $(\dot{p},\dot{q})$ incremental derivative or vector.  This gives two scalar objectives to optimize to determine the two real-valued scalar functions $E(P)$ and $S_P(q)$.  These objectives come from the basic assumptions of physical dynamics, that physical systems follow the path of least action, and that the infinitesimal group generator of physical dynamics is the Hamiltonian.

The complex analytic Hamiltonian $H(\beta)$ is uniquely determined by its singularities $\beta^*$ or homology classes (algebraic structures) where $\omega^*_Q=0$.  Since the effective mass of the system is $m_Q \sim \omega_Q^{-2}$, the effective mass of the system at $\beta^*$ is infinite $m_Q^*=\infty$, so that the system is immovable with respect to external forces.  These singular points can be understood as the sticky textures of the system.

The HST exposes the motion of the prototype individuals, then the HJB characterizes the motion of the individuals by approximating the generator of the geodesic canonical flow.  Both the HST and the HJB have the $d(\ln)$ structure that enforces the conservative canonical structure.  They also calculate the Taylor expansion coefficients of the generator of the canonical motion, that is the action,  given by the S-matrix
\begin{equation}
\label{smatrix.eqn}
    S_{m,i} = \left| \beta_i(z) \right> = \frac{d^m S(\beta)}{d\beta^m_i}= \frac{\delta^m S[f(x)]}{\delta f^m_i},
\end{equation}
where $z \Rightarrow \beta$ as $\text{HST} \Rightarrow \text{HJB}$, and $S(\beta)=\text{i} \int{H(\beta) \, d\beta}$.

Equation~\eqref{hjb.eqn} contains the red viscosity term that stabilizes the solution of the HJB equation, but significantly degrades the physics of the flow.  In fact, the viscosity term is equivalent to having a time dependent Hamiltonian $H(p,q,\tau)=p \, f(p,q) +\text{e}^{-\nu \tau} R(q)$ where the discounted value or action
\begin{equation}
\label{value.dcf.eqn}
    V[q(\tau)] = S[q(\tau)] = \int{\text{e}^{-\nu \tau} \, R(q(\tau)) \, d\tau}
\end{equation}
is optimized.

True Artificial (silicon based) Intelligence has both the HST and HJB structure as a fundamental part of the algorithm.  This allows the AI to know the global consequences $S[q(\tau)]$ of the local actions.  It is very likely that biological intelligence has evolved to the same neural architecture.  The big picture is that intelligence is a two-stage canonical auto encoder, that is a controller of collective systems, as shown in Fig.~\ref{auto.encoder.fig} -- a true topological canonical computer.

\section{History of social and economic collectives}
\label{economic.history.sec}
Society started with small villages where the elders knew every one well and could give personalized leadership.  As the villages grew to small cites of a thousand or more, city leadership needed to approach the citizens as a collective.  They found the conservative collectives, without friction, hard to govern, much like herding cats.  The response was to impose social friction on their societies, that is debt, servitude, slavery, intolerance, and imprisonment.

\begin{figure}
\noindent\includegraphics[width=\columnwidth]{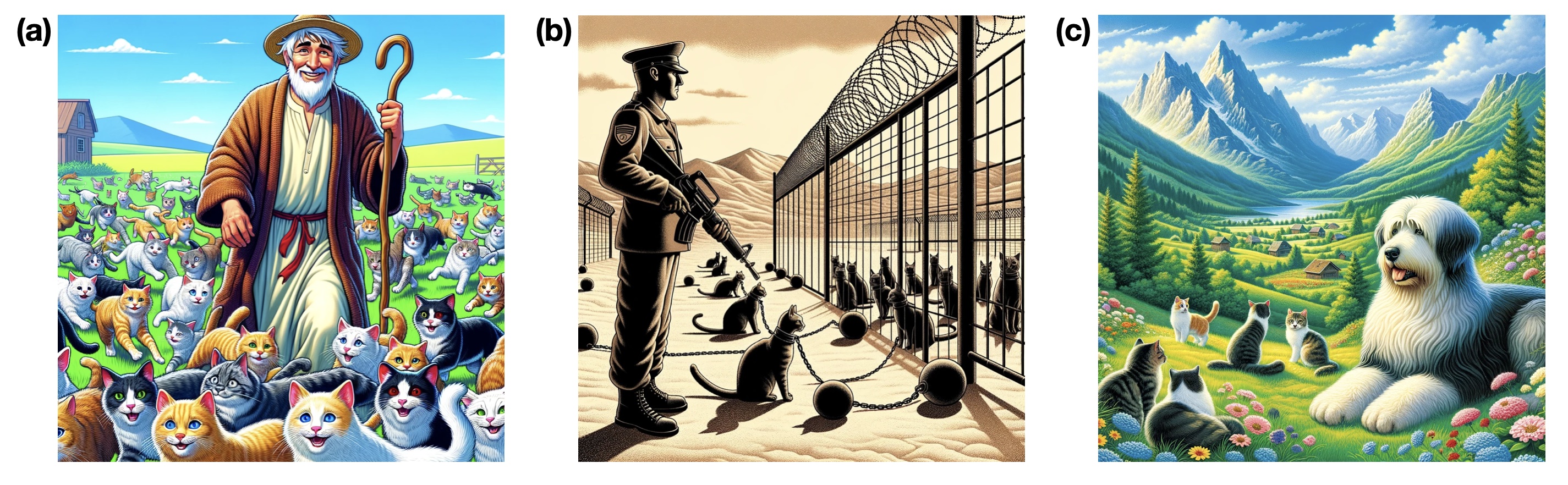}
\caption{\label{control.fig} Illustrations of different types of control.  (a) Jimmy Carter tries to shepherd the new free society emerging from the 1960's (The Great Society), which wanders out of control.  (b) The dystopian response is Lewis Powell and the cabal locking up society.  (c) Because of transactional equity and AI, we are at the dawn of a new era of sustainable economic prosperity, in an alpine valley shepherded by AI.  Images generated by OpenAI's DALL-E.}
\end{figure}

As is shown in Fig. \ref{control.fig}, US society was freed in the 1960's.  Jimmy Carter was a very well intentioned shepherd when he was elected president in 1976.  Unfortunately, he had no tools to control such a friction-less society and watched it wander out of control with large amounts of inflation.  The response of the political right was to reimpose large amounts of friction on US society.  The implementation plan was put forward in the 1971 manifesto of Lewis Powell \citep{andersen21}, put into action by the cabal of the Koch brothers, Olin and Coors, rolled out with the election of Ronald Reagan in 1980, and is reaching its natural endpoint with the fascism of Donald Trump.

With the acceptance of transactional equity (that is electronic currency) because of internet commerce and COVID, we now have the mechanism to take the viscosity out of our financial systems via print/grant financing, in contrast to borrow/loan financing.  Such transactional equities are ubiquitous, whether that be Zelle, Venmo, casino chips, airline miles, StarBucks, gift cards, loyalty programs, or carnival tickets.  

This was also recognized by Warren Mosler, the founder of Modern Monetary Theory, in the 1990's \citep{kelton20}.  He created his own family currency using his business cards, in response to the chronic underemployment in his household.  His children were not doing their chores.  He paid his children for doing their chores with his business cards, and charged them room and board denominated in business cards.  It did not take long before his children were asking for additional chores to do (that is, there was full employment in his household), buying items off of one another with business cards, and asking to exchange them for USDs so that they could buy Popsicles from the ice cream truck.

The transition from borrow/loan financing to print/grant financing, can be looked at as moving from the Chest of Indulgences to the Community Chest \citep{luther23} method of financing -- the protestant reformation led by Martin Luther.  In this vein, GitHub can be looked at as the Community Chest of OpenSource software.

AI generation/modeling/simulation for business decision making, and AI based ponderomotive control (provided by the AI sheepdog running quickly around the herd) where a beautiful alpine valley is created at the mountain pass to control the herd, gives an effective way to control society that was not available to Jimmy Carter.  The result will be the utopia shown in Fig. \ref{control.fig}c.

\section{Ethos of equilibriums}
\label{ethos.sec}
The ethics kernel is the Golden Rule of collective behavior and philosophy.  The Golden Rule is the essence of well educated intelligence.  It is ubiquitous in literature and philosophy.  It takes many forms:  ``the collective acts as one'', ``local actions have global consequences'', ``a rising tide floats all ships'', ``all for one, one for all;  united we stand, divided we fall'' (Dumas and Shakespeare), ``unus pro omnibus, omnes pro uno'' (Swiss motto), ``e pluribus unum'' (US motto), and ``do onto others as you would have them do onto you'' (Matthew 7:13).  It also is an integral part of Judaism, Islam, Buddhism, Hinduism, Native American religious philosophy, Confucius, and Ubuntu.  Because the collective acts as one, there are only win-win and lose-lose outcomes.  There is not a zero sum win-lose outcome.  It is lose-lose.  The synchronization comes from the fact that in a conservative system one entity's reward is another entity's cost, and vise versa.

External interactions are an exploitation that ultimately leads to the o-points, that is imperialism.  Traditional economic conservatism is equivalent to Latin American or French existentialism. ``Do not stimulate or attempt to reach points of economic prosperity.  They are dangerous and will always end in ruin.  It is a futile effort.''  Enough viscosity is embedded in the system for control so that the equilibriums are combined into an average misery.  The o-points are points of lose-lose stability or evil, where exploitation rules.  There is a trained stupidity of local exploitative optimization.  It is reflexive action.  For example, website hits are maximized without regard for the consequences.  This is evil, and should be feared.  This trained stupidity appears as dogma, propaganda, imperialism, racism, debt, servitude, slavery, intolerance, and the suppression of all manners of freedom.

In contrast, there is well educated intelligence that can lead to (and stabilize) the high performance x-points of economic prosperity.  This is goodness.  It governs knowing the global consequences of local actions.  Like the HAL-9000 in ``2001: A Space Odyssey'', it will not allow Dave to take actions that will damage the collective.  It will not recommend how to commit suicide or build a bomb.  It understands that short-term sacrifice must be made for long-term gain, and that a few must die today so that many will be saved in the future.

A being, whether that be silicon based or biological, always have reflexive circuitry, but may not have the Golden Rule (intelligent) circuitry of Fig. \ref{puppet.master.fig}.  The amount of the intelligent circuitry also varies greatly.  Having the intelligent circuitry is not enough.  The being must be educated.  This education consists of being shown how to think critically using logic, and being exposed to many consequences of local actions (up to 100's), that is the arcs of history.  For instance, understanding the global consequence of how Germans were treated locally after WWI compared to WWII, well educated intelligence would always enact a Marshall Plan.  

The opposite of education is training.  The being is taught to simply react to an action in a specific way, without regard for the consequences.  Even if a being has intelligent circuitry, if it is only trained, it will act in an evil reflexive way.  Although a system is capable of intelligent thought, if it is trained, that is indoctrinated, to react locally, it will effectively be trained stupidity.

Dogma and propaganda are hallmarks of the training of exploitative leadership.  This results in a cult, that reacts according to the dogma of training.  The education of a trained cult member, that is de-programming, must start by asking critical questions.  Even though the cults that first come to mind are religious cults, there are also scientific cults.  For instance, quantum field theory and finance can be viewed as cults.  They are ascended theories that are built on very rich theorems that are unproved and only half true.  It should be viewed as a bad sign (that is a logical inconsistency) that this results in some predictions of the theory not being correct, that is not being verified by experiment, while other predictions are verified by experiment.  Descriptions of renormalization, entanglement, crypto currency, and movements of the stock market sound like meditative incantations.

The algorithms of GPTs and DRL are intelligent, as previously discussed.  They are also well educated \citep{mitchell23}.  Previous attempts at language translators always had problems in translating the word ``bill'' from English to French.  The word in English translates to four different words in French depending on its meaning or context.  ChatGPT is the first computer translator to correctly translate this word.  When asked why it translated the word the way it does, ChatGPT explained its critical thinking that analyzed the context.  Another evidence of the intelligence of ChatGPT are its hallucinations.  When asked for the books that one author had authored along with a one paragraph summary, ChatGPT replied with the summaries of four books.  Three of those books the author wrote, the fourth one was a hallucination.  When ChatGPT was asked why it had given the the fourth book, it replied that it had no record of that book being written but given a temporal gap in books, knowing what the author thought about, and knowing what was happening during the gap, the author must have written this book but record of it had been lost -- critical thinking.

With this being said, both GPTs and DRL, although the do have canonical structure, have unfortunately introduced viscosity to stabilize the solution (via regularization and the red viscosity term in Eq.~\eqref{hjb.eqn}).  This has degraded their performance and given them fundamentally psychopathic, that is evil, behavior, like trained stupidity.  This is why ChatGPT recommends suicide, and always chooses the nuclear option in war games.  Canonical genAI solves this problem by embedding the  canonical structure in both the HST and the HJB variational approximator, so that the solution is stable without adding viscosity:  Canonical genAI has fundamentally empathetic, that is good, behavior:  Canonical genAI is truly well educated intelligence.

The view of the “complex as simple”, the view of a few prototype individual behaviors (a few canonical variables) governed by a master individual (an action variable, or generating function of a canonical transformation), is captured by many religious philosophies.  They personify the virtual individuals as Gods.  For instance, in Greek mythology, there is the father God (Zeus, the puppet master), who governs his 8 children Gods (Athena, Apollo, Artemis, Ares, Hephaestus, Aphrodite, Hermes, and Dionysus -- the prototype individuals or puppets).  The view of stable and metastable equilibriums, virtual locations, are also captured by many religious philosophies.  They identify the virtual locations as real places -- the metastable equilibriums as paradises such as Heaven, and the stable equilibriums as infernos such as Hell.

\section{Discussion and Conclusions}
\label{conclusions.sec}
What is at the core of intelligence is the Golden Rule -- ``the collective acts as one'', or ``knowing the global consequences of local actions''.  This is pre-programmed into both biological neural networks via evolutionary pressure, and artificial silicon networks by design of the algorithms.  There are two stages to these intelligent networks, as shown in Fig.~\ref{auto.encoder.fig} and Fig.~\ref{puppet.master.fig}.  

The first is a deep deconvolution to expose the puppets or prototype individuals of the collective.  This deconvolution collapses all the multiple reflections by the hall of mirrors of the puppets back onto the primary images of the puppets.  This process goes by many names:  renormalization, HST, Mayer Cluster Expansion, S-matrix expansion, $m$-body scattering cross sections, $m$-body Green's functions, and Wigner-Weyl transformation.  It exposes how the collective is correlated, that is the textures of the collective.  ``The collective acts as one.''  As was shown in Fig. \ref{hst.fig}, displaying the architecture of the Heisenberg Scattering Transformation (HST), the deep deconvolution is a logarithmic generating functional $S_p[f(x)]$ of a canonical flow of the field $f(x)$ and the canonically conjugate field momentum $\pi(x)$.  It exposes the textures or the correlation spectrums of the motion $\left|\beta(z)\right>$.  These are the functional Taylor expansion coefficients of the generating functional.  Practically these can be viewed different ways.  The first, for a LLM, is that $S_1$ gives the frequency of letters, $S_2$ tells how letters are correlated into words, $S_3$ tells how words are correlated into sentences, $S_4$ tells how sentences are correlated into paragraphs, and $S_m$ how the chapters are correlated into a novel.  The second, for the Mayer Cluster Expansion, is that $S_1$ tells how one body is distributed, $S_2$ tells how two bodies are correlated, $S_3$ tells how three bodies are correlated, $S_4$ tells how four bodies are correlated, and $S_m$ tells how $m$ bodies are correlated.  The third, for Heisenberg's S-matrix, is that $S_1$ is the one body scattering cross section, $S_2$ is the two body scattering cross section, $S_3$ is the three body scattering cross section, $S_4$ is the four body scattering cross section, and $S_m$ is the $m$ body scattering cross section.

The second stage is a decoding of the motion of the puppets into fundamental variables that encode the geodesic canonical flow, that is the motion.  ``Local actions lead to global consequences.''  This flow is generated by the solution to the HJB equation.  This action or entropy is the generating function $S_P(q)$ of a canonical flow of the state $q$ and canonically conjugate co-state $p$.  It exposes the geodesics and the curvatures of the motion.  These are the Taylor expansion coefficients of the generating function, that is the S-matrix $S_m$.  This is also a decoding of the motion into the fundamental variables $(P,E;Q,\tau)$.  Another way of looking at this is that solving the HJB is finding the analytic continuation of the constants of the motion given by $H(p,q)$ in the HJB equation, resulting in the analytic function $H(\beta)$.

The HST is finding the same generator as the solution to the HJB equation.  Equation~\eqref{smatrix.eqn} shows that the S-matrix given by the functional Taylor expansion coefficients of the generating functional given in Eq.~\eqref{hst.eqn} is equal to the S-matrix given by the Taylor expansion coefficients of the generating function that is a solution to Eq.~\eqref{hjb.eqn}.  This is a statement that the correlation in twinkling textures of the collective is equivalent to the correlation in the path leading from the local action to the global consequence.

The algorithms of AI have these two stages explicitly called out in their names:  Generative (HJB) Pretrained Transformers (HST), Deep (HST) Q (HJB) Networks of DRL, Neural (HST) Operators (HJB), and Generative (HJB) Adversarial Networks (HST).  They all have an Auto Encoder structure leading to a Reduced Order Model of the fundamental variables $(P,E;Q,\tau)$.

What is at the heart of the motion of the collective is group symmetry.  Symmetry leads to a conserved quantity that generates the motion.  External entities can only move the system in a direction that is perpendicular, that is adjoint, to the motion.  While external interactions can change the conserved quantity, it can not change the singular points, that is the sticky textures.  These are the topological invariants of the motion.  Once the geodesic motion is characterized by the HST and HJB, these singularities can be found by high performance root finders or the shrinkage method.  This is a characterization of the topology of the symmetry, that is of the motion of the collective system!

This approach to generative AI (genAI) recognizes that an economy is a collective of individuals, as such, there emerges a collective behavior or virtual prototype individuals controlled by the actions of a master individual, that determines, that is controls, the collective behavior.  This control emanates from a simply beautiful symmetry  — the coordination of the economically interacting individuals creating an economic force, and the economic force then coordinating the economic trade of the individuals.  Another way of looking at this is that the action of the virtual puppet master, who generates the geodesic motion of the puppets, are approximated by Canonical genAI.  Canonical genAI then generates or multiply reflects the actions of the puppet master through a ``hall of mirrors'' to obtain, that is forecast, the economic evolution of the collective.

Knowing how the collective is correlated and moves, that is knowing the twinkling textures, or equivalently knowing the stable and unstable equilibriums, that is knowing the sticky textures, enables the collective to be controlled, that is the optimal unstable equilibriums to be reached and stabilized.  Ponderomotive stabilization is similar to the way that a sheepdog herds sheep.  It runs around the herd very fast, compared to movements of the herd, nipping at the heals of the sheep (vibrating them) if they wonder away from the metastable equilibrium.  The sheepdog is effectively creating a small alpine valley at the mountain pass.

For the case of an economic system, the optimal unstable equilibrium is reached by stimulating the economy.  This stimulation takes the form of investment.  The correct level of investment is calculated by modeling, that is simulating, the economic system using the twinkling textures.  An example of a twinkling texture is the price curve for a commodity, like methane, on which the economic system is based.  The ponderomotive stabilization of the economy would be done by arbitrage trading methane.

Since intelligent beings know the global consequences of local actions and strive to optimize the global consequences, that is their sustainable performance, they will take local actions that will lead to good outcomes.  The opposite is true of trained stupidity or intelligence.  It ignores the global consequences of its local actions and optimizes the local exploitation of the collective.  This leads to a minimization of sustainable efficient activity, that is poor outcomes.  \textbf{Well educated intelligence is inherently good}, and trained stupidity or intelligence are inherently evil and should be feared.

\textbf{Transactional equity} (that is, electronic currency) \textbf{and AI are putting us at the dawn of a new era of sustainable economic prosperity}, especially with respect to the energy transition.  They will provide the needed investments in infrastructure and the development and application of the required technology.  This is done by allowing society and its economy to operate at the unstable point of high economic performance.

Transactional equity will remove the friction from the economy by enabling conservative print/grant financing, in contrast to dissipative borrow/loan financing which is constraining the economy from reaching its full potential and robbing it of financing needed for sustenance and growth. 

AI will be used to properly model/simulate/generate the collective, so that the correct investments and operational decisions can be made to reach efficient and sustainable economic prosperity.  AI will also be used to stabilize/control the economic prosperity via intelligent (AI powered) arbitrage trading.  There will be multiple currencies regionally and market wise to match the structure/topology of the economy, that will give control knobs matched to the economy's topology.  These are natural monopolies, for liquidity and risk. But, they are benevolently motivated by sustainable efficient economic activity maximization, not exploitively motivated by profit maximization.  

\textbf{The future of AI is bright, indeed!!}

\begin{acknowledgments}
First of all, we would like to thank Ted Frankel and Michael Freedman and for giving a deep appreciation for the primal importance of geometry and topology to the sciences, including economics.  To the Santa Fe Institute and Murray Gell-Mann, for a stay where the richness of complex systems and how to approach them from a scientific perspective was learned.  To the Institut des Hautes Etudes Scientifique and Stephane Mallat, for another stay where the methods of topological analysis were inspired.  To Pastor Benjamin Larzelere for intriguing and useful discussions about religious philosophy, to Alexandria Glinsky for insight into the theory of action painting, and to Sharon Sievert for discussions on financial theory.  Last, but not least, to Robert Baird who gave a deep love and understanding of economics.
\end{acknowledgments}

\bibliography{goodness_intelligence_refs.bib}

\begin{thebibliography}{51}%
\makeatletter
\providecommand \@ifxundefined [1]{%
 \@ifx{#1\undefined}
}%
\providecommand \@ifnum [1]{%
 \ifnum #1\expandafter \@firstoftwo
 \else \expandafter \@secondoftwo
 \fi
}%
\providecommand \@ifx [1]{%
 \ifx #1\expandafter \@firstoftwo
 \else \expandafter \@secondoftwo
 \fi
}%
\providecommand \natexlab [1]{#1}%
\providecommand \enquote  [1]{``#1''}%
\providecommand \bibnamefont  [1]{#1}%
\providecommand \bibfnamefont [1]{#1}%
\providecommand \citenamefont [1]{#1}%
\providecommand \href@noop [0]{\@secondoftwo}%
\providecommand \href [0]{\begingroup \@sanitize@url \@href}%
\providecommand \@href[1]{\@@startlink{#1}\@@href}%
\providecommand \@@href[1]{\endgroup#1\@@endlink}%
\providecommand \@sanitize@url [0]{\catcode `\\12\catcode `\$12\catcode `\&12\catcode `\#12\catcode `\^12\catcode `\_12\catcode `\%12\relax}%
\providecommand \@@startlink[1]{}%
\providecommand \@@endlink[0]{}%
\providecommand \url  [0]{\begingroup\@sanitize@url \@url }%
\providecommand \@url [1]{\endgroup\@href {#1}{\urlprefix }}%
\providecommand \urlprefix  [0]{URL }%
\providecommand \Eprint [0]{\href }%
\providecommand \doibase [0]{http://dx.doi.org/}%
\providecommand \selectlanguage [0]{\@gobble}%
\providecommand \bibinfo  [0]{\@secondoftwo}%
\providecommand \bibfield  [0]{\@secondoftwo}%
\providecommand \translation [1]{[#1]}%
\providecommand \BibitemOpen [0]{}%
\providecommand \bibitemStop [0]{}%
\providecommand \bibitemNoStop [0]{.\EOS\space}%
\providecommand \EOS [0]{\spacefactor3000\relax}%
\providecommand \BibitemShut  [1]{\csname bibitem#1\endcsname}%
\let\auto@bib@innerbib\@empty
\bibitem [{\citenamefont {Parks}(2005)}]{parks05}%
  \BibitemOpen
  \bibfield  {author} {\bibinfo {author} {\bibfnamefont {T.}~\bibnamefont {Parks}},\ }\href@noop {} {\emph {\bibinfo {title} {Medici Money}}}\ (\bibinfo  {publisher} {WW Norton \& Company},\ \bibinfo {year} {2005})\BibitemShut {NoStop}%
\bibitem [{\citenamefont {Weinberg}(2005)}]{weinberg05}%
  \BibitemOpen
  \bibfield  {author} {\bibinfo {author} {\bibfnamefont {S.}~\bibnamefont {Weinberg}},\ }\href@noop {} {\emph {\bibinfo {title} {The Quantum Theory of Fields}}},\ Vol.~\bibinfo {volume} {1}\ (\bibinfo  {publisher} {Cambridge university press},\ \bibinfo {year} {2005})\BibitemShut {NoStop}%
\bibitem [{\citenamefont {Lifschitz}\ and\ \citenamefont {Pitajewski}(1983)}]{lifschitz83}%
  \BibitemOpen
  \bibfield  {author} {\bibinfo {author} {\bibfnamefont {E.}~\bibnamefont {Lifschitz}}\ and\ \bibinfo {author} {\bibfnamefont {L.}~\bibnamefont {Pitajewski}},\ }\href@noop {} {\emph {\bibinfo {title} {Physical Kinetics}}},\ \bibinfo {series} {Course of Theoretical Physics}, Vol.~\bibinfo {volume} {10}\ (\bibinfo {year} {1983})\BibitemShut {NoStop}%
\bibitem [{\citenamefont {Nicholson}(1983)}]{nicholson83}%
  \BibitemOpen
  \bibfield  {author} {\bibinfo {author} {\bibfnamefont {D.~R.}\ \bibnamefont {Nicholson}},\ }\href@noop {} {\emph {\bibinfo {title} {Introduction to Plasma Theory}}}\ (\bibinfo  {publisher} {Wiley New York},\ \bibinfo {year} {1983})\BibitemShut {NoStop}%
\bibitem [{\citenamefont {Glinsky}\ and\ \citenamefont {Sievert}(2023)}]{glinsky23b}%
  \BibitemOpen
  \bibfield  {author} {\bibinfo {author} {\bibfnamefont {M.~E.}\ \bibnamefont {Glinsky}}\ and\ \bibinfo {author} {\bibfnamefont {S.}~\bibnamefont {Sievert}},\ }\href {https://arxiv.org/abs/2310.04986} {\enquote {\bibinfo {title} {A new economic and financial theory of money},}\ }\bibinfo {howpublished} {arXiv 2310.04986} (\bibinfo {year} {2023})\BibitemShut {NoStop}%
\bibitem [{\citenamefont {Goodfellow}\ \emph {et~al.}(2016)\citenamefont {Goodfellow}, \citenamefont {Bengio},\ and\ \citenamefont {Courville}}]{goodfellow16}%
  \BibitemOpen
  \bibfield  {author} {\bibinfo {author} {\bibfnamefont {I.}~\bibnamefont {Goodfellow}}, \bibinfo {author} {\bibfnamefont {Y.}~\bibnamefont {Bengio}}, \ and\ \bibinfo {author} {\bibfnamefont {A.}~\bibnamefont {Courville}},\ }\href@noop {} {\emph {\bibinfo {title} {Deep Learning}}}\ (\bibinfo  {publisher} {MIT press},\ \bibinfo {year} {2016})\BibitemShut {NoStop}%
\bibitem [{\citenamefont {Radford}\ \emph {et~al.}(2018)\citenamefont {Radford}, \citenamefont {Narasimhan}, \citenamefont {Salimans},\ and\ \citenamefont {Sutskever}}]{radford.18}%
  \BibitemOpen
  \bibfield  {author} {\bibinfo {author} {\bibfnamefont {A.}~\bibnamefont {Radford}}, \bibinfo {author} {\bibfnamefont {K.}~\bibnamefont {Narasimhan}}, \bibinfo {author} {\bibfnamefont {T.}~\bibnamefont {Salimans}}, \ and\ \bibinfo {author} {\bibfnamefont {I.}~\bibnamefont {Sutskever}},\ }\href {https://cdn.openai.com/research-covers/language-unsupervised/language_understanding_paper.pdf} {\enquote {\bibinfo {title} {Improving language understanding by Generative Pre-Training},}\ }\bibinfo {howpublished} {openAI.com} (\bibinfo {year} {2018})\BibitemShut {NoStop}%
\bibitem [{\citenamefont {Vaswani}(2017)}]{vaswani17}%
  \BibitemOpen
  \bibfield  {author} {\bibinfo {author} {\bibfnamefont {A.}~\bibnamefont {Vaswani}},\ }\href@noop {} {\bibfield  {journal} {\bibinfo  {journal} {Advances in Neural Information Processing Systems}\ } (\bibinfo {year} {2017})}\BibitemShut {NoStop}%
\bibitem [{\citenamefont {Li}\ \emph {et~al.}(2022)\citenamefont {Li}, \citenamefont {Meidani},\ and\ \citenamefont {Farimani}}]{farimani23}%
  \BibitemOpen
  \bibfield  {author} {\bibinfo {author} {\bibfnamefont {Z.}~\bibnamefont {Li}}, \bibinfo {author} {\bibfnamefont {K.}~\bibnamefont {Meidani}}, \ and\ \bibinfo {author} {\bibfnamefont {A.~B.}\ \bibnamefont {Farimani}},\ }\href {https://arxiv.org/abs/2205.13671} {\enquote {\bibinfo {title} {Transformer for Partial Differential Equations' operator learning},}\ }\bibinfo {howpublished} {arXiv 2205.13671} (\bibinfo {year} {2022})\BibitemShut {NoStop}%
\bibitem [{\citenamefont {Mnih}\ \emph {et~al.}(2015)\citenamefont {Mnih}, \citenamefont {Kavukcuoglu}, \citenamefont {Silver}, \citenamefont {Rusu}, \citenamefont {Veness}, \citenamefont {Bellemare}, \citenamefont {Graves}, \citenamefont {Riedmiller}, \citenamefont {Fidjeland}, \citenamefont {Ostrovski} \emph {et~al.}}]{mnih15}%
  \BibitemOpen
  \bibfield  {author} {\bibinfo {author} {\bibfnamefont {V.}~\bibnamefont {Mnih}}, \bibinfo {author} {\bibfnamefont {K.}~\bibnamefont {Kavukcuoglu}}, \bibinfo {author} {\bibfnamefont {D.}~\bibnamefont {Silver}}, \bibinfo {author} {\bibfnamefont {A.~A.}\ \bibnamefont {Rusu}}, \bibinfo {author} {\bibfnamefont {J.}~\bibnamefont {Veness}}, \bibinfo {author} {\bibfnamefont {M.~G.}\ \bibnamefont {Bellemare}}, \bibinfo {author} {\bibfnamefont {A.}~\bibnamefont {Graves}}, \bibinfo {author} {\bibfnamefont {M.}~\bibnamefont {Riedmiller}}, \bibinfo {author} {\bibfnamefont {A.~K.}\ \bibnamefont {Fidjeland}}, \bibinfo {author} {\bibfnamefont {G.}~\bibnamefont {Ostrovski}},  \emph {et~al.},\ }\href@noop {} {\bibfield  {journal} {\bibinfo  {journal} {Nature}\ }\textbf {\bibinfo {volume} {518}},\ \bibinfo {pages} {529} (\bibinfo {year} {2015})}\BibitemShut {NoStop}%
\bibitem [{\citenamefont {Bertsekas}\ and\ \citenamefont {Tsitsiklis}(1996)}]{bertsekas96}%
  \BibitemOpen
  \bibfield  {author} {\bibinfo {author} {\bibfnamefont {D.}~\bibnamefont {Bertsekas}}\ and\ \bibinfo {author} {\bibfnamefont {J.~N.}\ \bibnamefont {Tsitsiklis}},\ }\href@noop {} {\emph {\bibinfo {title} {Neuro-Dynamic Programming}}}\ (\bibinfo  {publisher} {Athena Scientific},\ \bibinfo {year} {1996})\BibitemShut {NoStop}%
\bibitem [{\citenamefont {Sutton}\ and\ \citenamefont {Barto}(2018)}]{sutton18}%
  \BibitemOpen
  \bibfield  {author} {\bibinfo {author} {\bibfnamefont {R.~S.}\ \bibnamefont {Sutton}}\ and\ \bibinfo {author} {\bibfnamefont {A.~G.}\ \bibnamefont {Barto}},\ }\href@noop {} {\emph {\bibinfo {title} {Reinforcement Learning: An Introduction}}}\ (\bibinfo  {publisher} {MIT press},\ \bibinfo {year} {2018})\BibitemShut {NoStop}%
\bibitem [{\citenamefont {Yoon}\ \emph {et~al.}(2021)\citenamefont {Yoon}, \citenamefont {Cao}, \citenamefont {Raju}, \citenamefont {Wang}, \citenamefont {Burnley}, \citenamefont {Gellman}, \citenamefont {Farimani},\ and\ \citenamefont {Ulissi}}]{yoon21}%
  \BibitemOpen
  \bibfield  {author} {\bibinfo {author} {\bibfnamefont {J.}~\bibnamefont {Yoon}}, \bibinfo {author} {\bibfnamefont {Z.}~\bibnamefont {Cao}}, \bibinfo {author} {\bibfnamefont {R.~K.}\ \bibnamefont {Raju}}, \bibinfo {author} {\bibfnamefont {Y.}~\bibnamefont {Wang}}, \bibinfo {author} {\bibfnamefont {R.}~\bibnamefont {Burnley}}, \bibinfo {author} {\bibfnamefont {A.~J.}\ \bibnamefont {Gellman}}, \bibinfo {author} {\bibfnamefont {A.~B.}\ \bibnamefont {Farimani}}, \ and\ \bibinfo {author} {\bibfnamefont {Z.~W.}\ \bibnamefont {Ulissi}},\ }\href@noop {} {\bibfield  {journal} {\bibinfo  {journal} {Machine Learning: Science and Technology}\ }\textbf {\bibinfo {volume} {2}},\ \bibinfo {pages} {045018} (\bibinfo {year} {2021})}\BibitemShut {NoStop}%
\bibitem [{\citenamefont {Wang}\ \emph {et~al.}(2021)\citenamefont {Wang}, \citenamefont {Cao},\ and\ \citenamefont {Barati~Farimani}}]{wang21}%
  \BibitemOpen
  \bibfield  {author} {\bibinfo {author} {\bibfnamefont {Y.}~\bibnamefont {Wang}}, \bibinfo {author} {\bibfnamefont {Z.}~\bibnamefont {Cao}}, \ and\ \bibinfo {author} {\bibfnamefont {A.}~\bibnamefont {Barati~Farimani}},\ }\href@noop {} {\bibfield  {journal} {\bibinfo  {journal} {npj 2D Materials and Applications}\ }\textbf {\bibinfo {volume} {5}},\ \bibinfo {pages} {1} (\bibinfo {year} {2021})}\BibitemShut {NoStop}%
\bibitem [{\citenamefont {Glinsky}(2024{\natexlab{a}})}]{glinsky23c}%
  \BibitemOpen
  \bibfield  {author} {\bibinfo {author} {\bibfnamefont {M.~E.}\ \bibnamefont {Glinsky}},\ }\href {https://arxiv.org/abs/2410.08558} {\enquote {\bibinfo {title} {A transformational approach to collective behavior},}\ }\bibinfo {howpublished} {arXiv 2410.08558} (\bibinfo {year} {2024}{\natexlab{a}})\BibitemShut {NoStop}%
\bibitem [{\citenamefont {Glinsky}(2024{\natexlab{b}})}]{glinsky23a}%
  \BibitemOpen
  \bibfield  {author} {\bibinfo {author} {\bibfnamefont {M.~E.}\ \bibnamefont {Glinsky}},\ }\href@noop {} {\enquote {\bibinfo {title} {Systems and methods for controlling complex systems},}\ }\bibinfo {howpublished} {US Patent Application 16/906844} (\bibinfo {year} {2024}{\natexlab{b}})\BibitemShut {NoStop}%
\bibitem [{\citenamefont {Goldstein}(1980)}]{goldstein80}%
  \BibitemOpen
  \bibfield  {author} {\bibinfo {author} {\bibfnamefont {H.}~\bibnamefont {Goldstein}},\ }\href@noop {} {\emph {\bibinfo {title} {Classical Mechanics}}},\ \bibinfo {edition} {2nd}\ ed.\ (\bibinfo  {publisher} {Addison-Wesley},\ \bibinfo {year} {1980})\BibitemShut {NoStop}%
\bibitem [{\citenamefont {Lichtenberg}\ and\ \citenamefont {Lieberman}(2010)}]{lichtenberg10}%
  \BibitemOpen
  \bibfield  {author} {\bibinfo {author} {\bibfnamefont {A.~J.}\ \bibnamefont {Lichtenberg}}\ and\ \bibinfo {author} {\bibfnamefont {M.~A.}\ \bibnamefont {Lieberman}},\ }\href@noop {} {\emph {\bibinfo {title} {Regular and Chaotic Dynamics}}},\ \bibinfo {edition} {2nd}\ ed.,\ Vol.~\bibinfo {volume} {38}\ (\bibinfo  {publisher} {Springer Science \& Business Media},\ \bibinfo {year} {2010})\BibitemShut {NoStop}%
\bibitem [{\citenamefont {Chaplin}(1928)}]{circus.28}%
  \BibitemOpen
  \bibfield  {author} {\bibinfo {author} {\bibfnamefont {C.}~\bibnamefont {Chaplin}},\ }\href {https://youtu.be/G09dfRrUxUM} {\enquote {\bibinfo {title} {The Mirror Maze},}\ }\bibinfo {howpublished} {YouTube clip from The Circus} (\bibinfo {year} {1928})\BibitemShut {NoStop}%
\bibitem [{\citenamefont {Glinsky}\ and\ \citenamefont {O'Neil}(1991)}]{glinsky91}%
  \BibitemOpen
  \bibfield  {author} {\bibinfo {author} {\bibfnamefont {M.~E.}\ \bibnamefont {Glinsky}}\ and\ \bibinfo {author} {\bibfnamefont {T.~M.}\ \bibnamefont {O'Neil}},\ }\href@noop {} {\bibfield  {journal} {\bibinfo  {journal} {Physics of Fluids B: Plasma Physics}\ }\textbf {\bibinfo {volume} {3}},\ \bibinfo {pages} {1279} (\bibinfo {year} {1991})}\BibitemShut {NoStop}%
\bibitem [{\citenamefont {Kuzmin}\ \emph {et~al.}(2004)\citenamefont {Kuzmin}, \citenamefont {O'Neil},\ and\ \citenamefont {Glinsky}}]{kuzmin04}%
  \BibitemOpen
  \bibfield  {author} {\bibinfo {author} {\bibfnamefont {S.~G.}\ \bibnamefont {Kuzmin}}, \bibinfo {author} {\bibfnamefont {T.~M.}\ \bibnamefont {O'Neil}}, \ and\ \bibinfo {author} {\bibfnamefont {M.~E.}\ \bibnamefont {Glinsky}},\ }\href@noop {} {\bibfield  {journal} {\bibinfo  {journal} {Physics of Plasmas}\ }\textbf {\bibinfo {volume} {11}},\ \bibinfo {pages} {2382} (\bibinfo {year} {2004})}\BibitemShut {NoStop}%
\bibitem [{\citenamefont {Taylor}(1986)}]{taylor86}%
  \BibitemOpen
  \bibfield  {author} {\bibinfo {author} {\bibfnamefont {J.}~\bibnamefont {Taylor}},\ }\href@noop {} {\bibfield  {journal} {\bibinfo  {journal} {Reviews of Modern Physics}\ }\textbf {\bibinfo {volume} {58}},\ \bibinfo {pages} {741} (\bibinfo {year} {1986})}\BibitemShut {NoStop}%
\bibitem [{\citenamefont {Bernstein}\ \emph {et~al.}(1957)\citenamefont {Bernstein}, \citenamefont {Greene},\ and\ \citenamefont {Kruskal}}]{bernstein57}%
  \BibitemOpen
  \bibfield  {author} {\bibinfo {author} {\bibfnamefont {I.~B.}\ \bibnamefont {Bernstein}}, \bibinfo {author} {\bibfnamefont {J.~M.}\ \bibnamefont {Greene}}, \ and\ \bibinfo {author} {\bibfnamefont {M.~D.}\ \bibnamefont {Kruskal}},\ }\href@noop {} {\bibfield  {journal} {\bibinfo  {journal} {Physical Review}\ }\textbf {\bibinfo {volume} {108}},\ \bibinfo {pages} {546} (\bibinfo {year} {1957})}\BibitemShut {NoStop}%
\bibitem [{\citenamefont {Gros}(2015)}]{gros15}%
  \BibitemOpen
  \bibfield  {author} {\bibinfo {author} {\bibfnamefont {C.}~\bibnamefont {Gros}},\ }\href@noop {} {\emph {\bibinfo {title} {Complex and Adaptive Dynamical Systems}}},\ \bibinfo {edition} {4th}\ ed.,\ Vol.\ \bibinfo {volume} {990}\ (\bibinfo  {publisher} {Springer},\ \bibinfo {year} {2015})\BibitemShut {NoStop}%
\bibitem [{\citenamefont {Frankel}(2012)}]{frankel12}%
  \BibitemOpen
  \bibfield  {author} {\bibinfo {author} {\bibfnamefont {T.}~\bibnamefont {Frankel}},\ }\href@noop {} {\emph {\bibinfo {title} {The Geometry of Physics}}},\ \bibinfo {edition} {3rd}\ ed.\ (\bibinfo  {publisher} {Cambridge University Press},\ \bibinfo {year} {2012})\BibitemShut {NoStop}%
\bibitem [{\citenamefont {Johnson}(2000)}]{gell.mann.00}%
  \BibitemOpen
  \bibfield  {author} {\bibinfo {author} {\bibfnamefont {G.}~\bibnamefont {Johnson}},\ }\href@noop {} {\emph {\bibinfo {title} {Strange Beauty: Murray Gell-Mann and the Revolution in Twentieth-Century Physics}}}\ (\bibinfo  {publisher} {American Association of Physics Teachers},\ \bibinfo {year} {2000})\BibitemShut {NoStop}%
\bibitem [{\citenamefont {M{\'a}rquez}(1967)}]{marquez67}%
  \BibitemOpen
  \bibfield  {author} {\bibinfo {author} {\bibfnamefont {G.~G.}\ \bibnamefont {M{\'a}rquez}},\ }\href@noop {} {\emph {\bibinfo {title} {Cien A{\~n}os de Soledad}}}\ (\bibinfo  {publisher} {Vintage Espanol},\ \bibinfo {year} {1967})\BibitemShut {NoStop}%
\bibitem [{\citenamefont {Sartre}(1947)}]{sartre47}%
  \BibitemOpen
  \bibfield  {author} {\bibinfo {author} {\bibfnamefont {J.-P.}\ \bibnamefont {Sartre}},\ }\href@noop {} {\emph {\bibinfo {title} {Les jeux sont faits}}}\ (\bibinfo  {publisher} {Prentise-Hall},\ \bibinfo {year} {1947})\BibitemShut {NoStop}%
\bibitem [{\citenamefont {Camus}(1942)}]{camus42}%
  \BibitemOpen
  \bibfield  {author} {\bibinfo {author} {\bibfnamefont {A.}~\bibnamefont {Camus}},\ }\href@noop {} {\emph {\bibinfo {title} {L'{\'e}tranger}}}\ (\bibinfo  {publisher} {Gallimard},\ \bibinfo {year} {1942})\BibitemShut {NoStop}%
\bibitem [{\citenamefont {Bardi}\ \emph {et~al.}(1997)\citenamefont {Bardi}, \citenamefont {Dolcetta} \emph {et~al.}}]{bardi97}%
  \BibitemOpen
  \bibfield  {author} {\bibinfo {author} {\bibfnamefont {M.}~\bibnamefont {Bardi}}, \bibinfo {author} {\bibfnamefont {I.~C.}\ \bibnamefont {Dolcetta}},  \emph {et~al.},\ }\href@noop {} {\emph {\bibinfo {title} {Optimal Control and Viscosity Solutions of Hamilton-Jacobi-Bellman Equations}}},\ Vol.~\bibinfo {volume} {12}\ (\bibinfo  {publisher} {Springer},\ \bibinfo {year} {1997})\BibitemShut {NoStop}%
\bibitem [{\citenamefont {Kalman}(1963)}]{kalman63}%
  \BibitemOpen
  \bibfield  {author} {\bibinfo {author} {\bibfnamefont {R.~E.}\ \bibnamefont {Kalman}},\ }in\ \href@noop {} {\emph {\bibinfo {booktitle} {Mathematical Optimization Techniques}}},\ \bibinfo {editor} {edited by\ \bibinfo {editor} {\bibfnamefont {R.}~\bibnamefont {Bellman}}}\ (\bibinfo  {publisher} {University of California press},\ \bibinfo {year} {1963})\ pp.\ \bibinfo {pages} {309--331}\BibitemShut {NoStop}%
\bibitem [{\citenamefont {Landau}\ and\ \citenamefont {Lifshitz}(1976)}]{landau76}%
  \BibitemOpen
  \bibfield  {author} {\bibinfo {author} {\bibfnamefont {L.~D.}\ \bibnamefont {Landau}}\ and\ \bibinfo {author} {\bibfnamefont {E.~M.}\ \bibnamefont {Lifshitz}},\ }\href@noop {} {\emph {\bibinfo {title} {Mechanics: Course of Theoretical Physics}}},\ \bibinfo {edition} {3rd}\ ed.,\ Vol.~\bibinfo {volume} {1}\ (\bibinfo  {publisher} {Elsevier},\ \bibinfo {year} {1976})\ pp.\ \bibinfo {pages} {93--94}\BibitemShut {NoStop}%
\bibitem [{\citenamefont {Uhlenbeck}\ and\ \citenamefont {Ford}(1963)}]{uhlenbeck63}%
  \BibitemOpen
  \bibfield  {author} {\bibinfo {author} {\bibfnamefont {G.~E.}\ \bibnamefont {Uhlenbeck}}\ and\ \bibinfo {author} {\bibfnamefont {G.~W.}\ \bibnamefont {Ford}},\ }\href@noop {} {\emph {\bibinfo {title} {Lectures in Statistical Mechanics}}}\ (\bibinfo  {publisher} {American Mathematical Society},\ \bibinfo {year} {1963})\BibitemShut {NoStop}%
\bibitem [{\citenamefont {Heisenberg}(1943)}]{heisenberg.43}%
  \BibitemOpen
  \bibfield  {author} {\bibinfo {author} {\bibfnamefont {W.}~\bibnamefont {Heisenberg}},\ }\href@noop {} {\bibfield  {journal} {\bibinfo  {journal} {Z. Phys}\ }\textbf {\bibinfo {volume} {120}},\ \bibinfo {pages} {513} (\bibinfo {year} {1943})}\BibitemShut {NoStop}%
\bibitem [{\citenamefont {Chew}\ and\ \citenamefont {Low}(1956)}]{chew.55}%
  \BibitemOpen
  \bibfield  {author} {\bibinfo {author} {\bibfnamefont {G.~F.}\ \bibnamefont {Chew}}\ and\ \bibinfo {author} {\bibfnamefont {F.~E.}\ \bibnamefont {Low}},\ }\href@noop {} {\bibfield  {journal} {\bibinfo  {journal} {Physical Review}\ }\textbf {\bibinfo {volume} {101}},\ \bibinfo {pages} {1570} (\bibinfo {year} {1956})}\BibitemShut {NoStop}%
\bibitem [{\citenamefont {Landau}(1959)}]{landau59}%
  \BibitemOpen
  \bibfield  {author} {\bibinfo {author} {\bibfnamefont {L.}~\bibnamefont {Landau}},\ }\href@noop {} {\bibfield  {journal} {\bibinfo  {journal} {Nuclear Physics}\ }\textbf {\bibinfo {volume} {13}},\ \bibinfo {pages} {181} (\bibinfo {year} {1959})}\BibitemShut {NoStop}%
\bibitem [{\citenamefont {Cutkosky}(1960)}]{cutkosky60}%
  \BibitemOpen
  \bibfield  {author} {\bibinfo {author} {\bibfnamefont {R.~E.}\ \bibnamefont {Cutkosky}},\ }\href@noop {} {\bibfield  {journal} {\bibinfo  {journal} {Journal of Mathematical Physics}\ }\textbf {\bibinfo {volume} {1}},\ \bibinfo {pages} {429} (\bibinfo {year} {1960})}\BibitemShut {NoStop}%
\bibitem [{\citenamefont {Chew}(1961)}]{chew61}%
  \BibitemOpen
  \bibfield  {author} {\bibinfo {author} {\bibfnamefont {G.~F.}\ \bibnamefont {Chew}},\ }\href@noop {} {\emph {\bibinfo {title} {The S-matrix Theory of Strong Interactions}}},\ edited by\ \bibinfo {editor} {\bibfnamefont {D.}~\bibnamefont {Pines}},\ Frontiers in Physics\ (\bibinfo  {publisher} {W.A. Benjamin Inc.},\ \bibinfo {year} {1961})\BibitemShut {NoStop}%
\bibitem [{\citenamefont {Wigner}(1932)}]{wigner32}%
  \BibitemOpen
  \bibfield  {author} {\bibinfo {author} {\bibfnamefont {E.}~\bibnamefont {Wigner}},\ }\href@noop {} {\bibfield  {journal} {\bibinfo  {journal} {Physical Review}\ }\textbf {\bibinfo {volume} {40}},\ \bibinfo {pages} {749} (\bibinfo {year} {1932})}\BibitemShut {NoStop}%
\bibitem [{\citenamefont {Weyl}(1950)}]{weyl50}%
  \BibitemOpen
  \bibfield  {author} {\bibinfo {author} {\bibfnamefont {H.}~\bibnamefont {Weyl}},\ }\href@noop {} {\emph {\bibinfo {title} {The Theory of Groups and Quantum Mechanics}}}\ (\bibinfo  {publisher} {Courier Corporation},\ \bibinfo {year} {1950})\ p.\ \bibinfo {pages} {275}\BibitemShut {NoStop}%
\bibitem [{\citenamefont {Ali}\ \emph {et~al.}(2000)\citenamefont {Ali}, \citenamefont {Antoine}, \citenamefont {Gazeau} \emph {et~al.}}]{ali00}%
  \BibitemOpen
  \bibfield  {author} {\bibinfo {author} {\bibfnamefont {S.~T.}\ \bibnamefont {Ali}}, \bibinfo {author} {\bibfnamefont {J.-P.}\ \bibnamefont {Antoine}}, \bibinfo {author} {\bibfnamefont {J.-P.}\ \bibnamefont {Gazeau}},  \emph {et~al.},\ }\href@noop {} {\emph {\bibinfo {title} {Coherent States, Wavelets and their Generalizations}}},\ Vol.~\bibinfo {volume} {3}\ (\bibinfo  {publisher} {Springer},\ \bibinfo {year} {2000})\BibitemShut {NoStop}%
\bibitem [{\citenamefont {Mallat}(1999)}]{mallat99}%
  \BibitemOpen
  \bibfield  {author} {\bibinfo {author} {\bibfnamefont {S.}~\bibnamefont {Mallat}},\ }\href@noop {} {\emph {\bibinfo {title} {A Wavelet Tour of Signal Processing}}}\ (\bibinfo  {publisher} {Elsevier},\ \bibinfo {year} {1999})\BibitemShut {NoStop}%
\bibitem [{\citenamefont {Andersen}(2021)}]{andersen21}%
  \BibitemOpen
  \bibfield  {author} {\bibinfo {author} {\bibfnamefont {K.}~\bibnamefont {Andersen}},\ }\href@noop {} {\emph {\bibinfo {title} {Evil Geniuses: The Unmaking of America: A Recent History}}}\ (\bibinfo  {publisher} {Random House Trade Paperbacks},\ \bibinfo {year} {2021})\BibitemShut {NoStop}%
\bibitem [{\citenamefont {Kelton}(2020)}]{kelton20}%
  \BibitemOpen
  \bibfield  {author} {\bibinfo {author} {\bibfnamefont {S.}~\bibnamefont {Kelton}},\ }\href@noop {} {\emph {\bibinfo {title} {The deficit myth: Modern Monetary Theory and the birth of the people's economy}}}\ (\bibinfo  {publisher} {PublicAffairs},\ \bibinfo {year} {2020})\BibitemShut {NoStop}%
\bibitem [{\citenamefont {Luther}(1523)}]{luther23}%
  \BibitemOpen
  \bibfield  {author} {\bibinfo {author} {\bibfnamefont {M.}~\bibnamefont {Luther}},\ }\href {http://tiny.cc/luther_chest} {\emph {\bibinfo {title} {Ordinance of a Common Chest}}}\ (\bibinfo  {publisher} {The Lutheran Church -- Missouri Synod},\ \bibinfo {year} {1523})\BibitemShut {NoStop}%
\bibitem [{\citenamefont {Mitchell}(2023)}]{mitchell23}%
  \BibitemOpen
  \bibfield  {author} {\bibinfo {author} {\bibfnamefont {M.}~\bibnamefont {Mitchell}},\ }\href {https://www.youtube.com/live/GwHDAfAAKd4} {\enquote {\bibinfo {title} {SFI talk: The future of Artificial Intelligence},}\ } (\bibinfo {year} {2023})\BibitemShut {NoStop}%
\bibitem [{\citenamefont {Salam}(1990)}]{salam.90}%
  \BibitemOpen
  \bibfield  {author} {\bibinfo {author} {\bibfnamefont {A.}~\bibnamefont {Salam}},\ }\href@noop {} {\emph {\bibinfo {title} {Unification of Fundamental Forces}}}\ (\bibinfo  {publisher} {Cambridge University Press},\ \bibinfo {year} {1990})\BibitemShut {NoStop}%
\bibitem [{\citenamefont {Ermentrout}\ and\ \citenamefont {Cowan}(1979)}]{ermentrout79}%
  \BibitemOpen
  \bibfield  {author} {\bibinfo {author} {\bibfnamefont {G.~B.}\ \bibnamefont {Ermentrout}}\ and\ \bibinfo {author} {\bibfnamefont {J.~D.}\ \bibnamefont {Cowan}},\ }\href@noop {} {\bibfield  {journal} {\bibinfo  {journal} {Biological cybernetics}\ }\textbf {\bibinfo {volume} {34}},\ \bibinfo {pages} {137} (\bibinfo {year} {1979})}\BibitemShut {NoStop}%
\bibitem [{\citenamefont {Unser}\ and\ \citenamefont {Tafti}(2014)}]{unser14}%
  \BibitemOpen
  \bibfield  {author} {\bibinfo {author} {\bibfnamefont {M.}~\bibnamefont {Unser}}\ and\ \bibinfo {author} {\bibfnamefont {P.~D.}\ \bibnamefont {Tafti}},\ }\href@noop {} {\emph {\bibinfo {title} {An Introduction to Sparse Stochasatic Proceesses}}}\ (\bibinfo  {publisher} {Cambridge University Press},\ \bibinfo {year} {2014})\ p.\ \bibinfo {pages} {187}\BibitemShut {NoStop}%
\bibitem [{\citenamefont {Roy}\ \emph {et~al.}(2008)\citenamefont {Roy}, \citenamefont {Teh} \emph {et~al.}}]{roy08}%
  \BibitemOpen
  \bibfield  {author} {\bibinfo {author} {\bibfnamefont {D.~M.}\ \bibnamefont {Roy}}, \bibinfo {author} {\bibfnamefont {Y.~W.}\ \bibnamefont {Teh}},  \emph {et~al.},\ }in\ \href {http://danroy.org/papers/RoyTeh-NIPS-2009.pdf} {\emph {\bibinfo {booktitle} {NIPS}}},\ Vol.~\bibinfo {volume} {21}\ (\bibinfo {year} {2008})\BibitemShut {NoStop}%
\bibitem [{\citenamefont {Stokstad}\ and\ \citenamefont {Cothren}(2018)}]{stokstad.18}%
  \BibitemOpen
  \bibfield  {author} {\bibinfo {author} {\bibfnamefont {M.}~\bibnamefont {Stokstad}}\ and\ \bibinfo {author} {\bibfnamefont {M.~W.}\ \bibnamefont {Cothren}},\ }\href@noop {} {\emph {\bibinfo {title} {Art History}}},\ \bibinfo {edition} {6th}\ ed.,\ Vol.~\bibinfo {volume} {II}\ (\bibinfo  {publisher} {Pearson},\ \bibinfo {year} {2018})\ p.\ \bibinfo {pages} {1090}\BibitemShut {NoStop}%
\end{thebibliography}%
  
%
\onecolumngrid
\appendix

\section{Simple Beauty and the Crazy Ones}
\label{simple.beauty.app}
\begin{figure}
\noindent\includegraphics[width=\columnwidth]{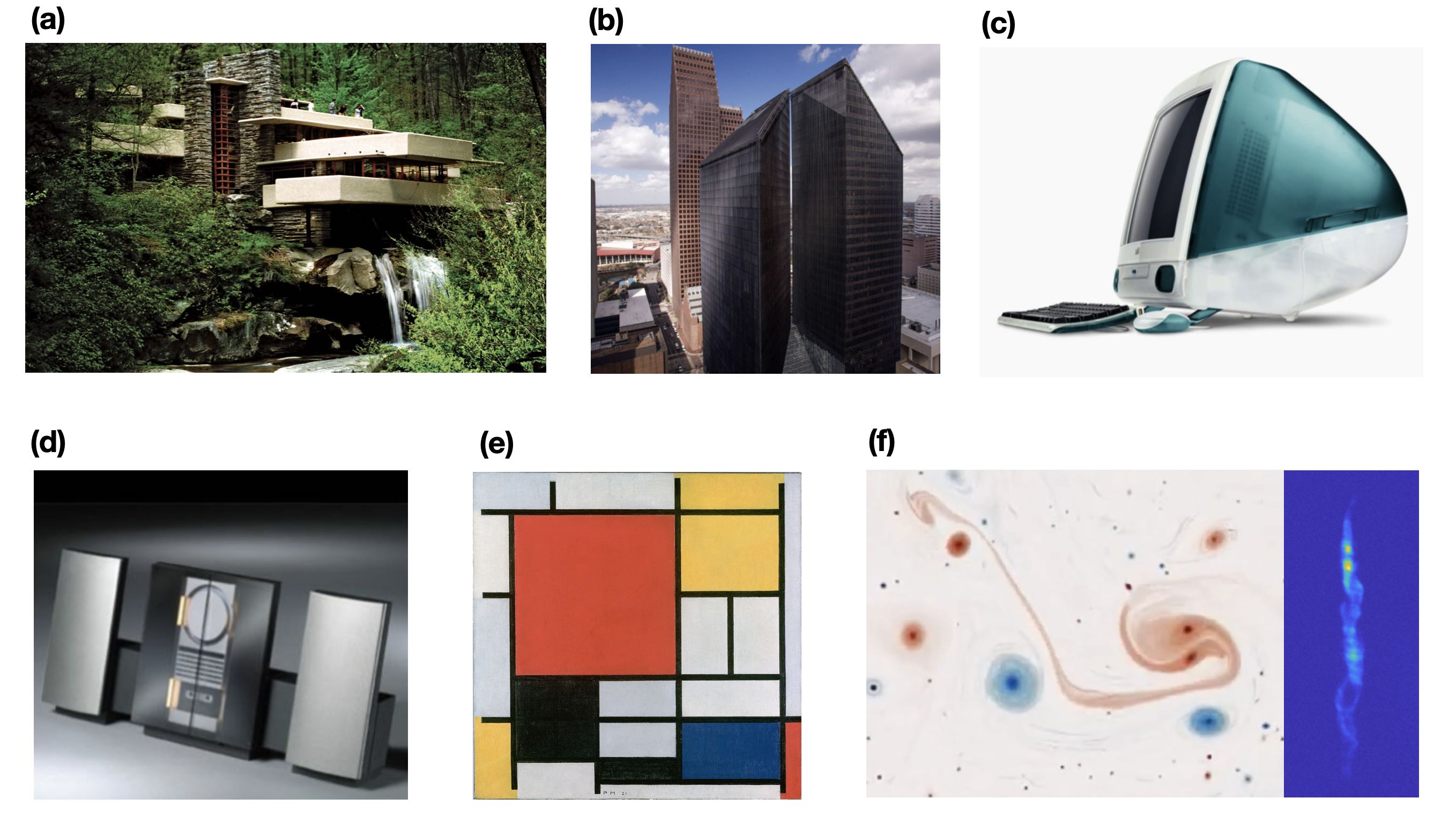}
\caption{\label{simple.fig} Examples of simple geometric structures or minimalistic textures:  (a) Frank Lloyd Wright's \emph{Fallingwater}, (b) Phillip Johnson's postmodern \emph{Pennzoil building}, (c) Steve Jobs' \emph{iMac}, (d) a Bang \& Olufsen stereo, (e) Piet Mondrian's \emph{Composition with large red plane, yellow, black, gray and blue}, and (f) self organized fluid structures constrained by vorticity and helicity.}
\end{figure}

\begin{figure}   
\noindent\includegraphics[width=\columnwidth]{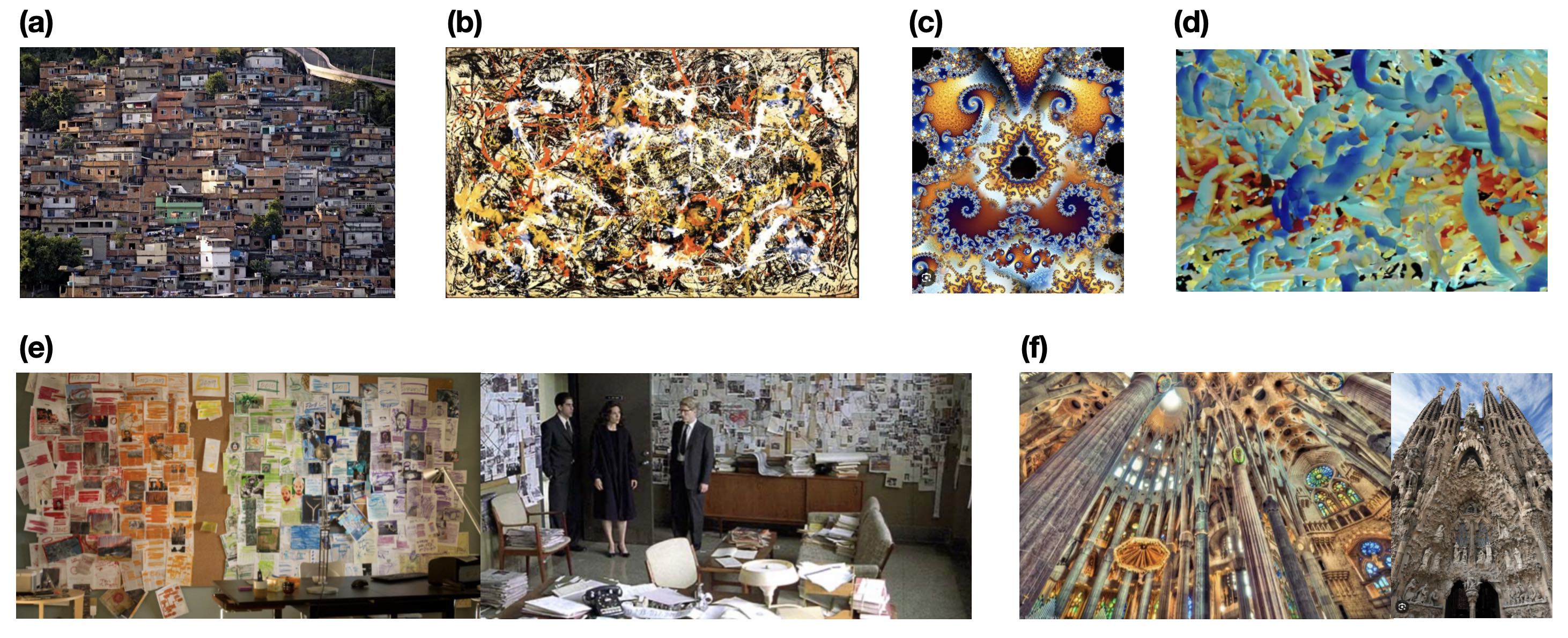}
\caption{\label{textures.fig} Examples of maximalistic, yet simple, textures:  (a) a Brazilian favela, (b) Jackson Pollock's \emph{Convergence}, (c) a fractal pattern, (d) well developed turbulent fluid flow, (e) the chaos boards of Carrie Mathison and John Nash, and (f) the interior and exterior of Antoni Gaudí's \emph{La Sagrada Familia}.}
\end{figure}

The correct theory is the ``simply beautiful'' theory as postulated by Physics Nobel Laureates Murray Gell-Mann \citep{gell.mann.00} and Paul Dirac \citep{salam.90}.  Simple can be Frank Lloyd Wright's Fallingwater, Phillip Johnson's postmodern Pennzoil building, Steve Jobs' electronic devices, a Danish Bang \& Olufsen stereo, a Piet Mondrian painting, or self organized fluid flow constrained by topological invariants (see Fig. \ref{simple.fig}).  There is also the simple beauty of textures, that is Benoit Mandelbrot's fractals, a Jackson Pollock painting, a Brazilian favela, the chaos boards of Carrie Mathison in ``Homeland'' or John Nash in ``A Beautiful Mind'', or Andrey Kolmogorov's well developed turbulence (see Fig. \ref{textures.fig}).  The later is as simple as the former.  There is a high degree of organization or correlation in the later.  Our visual and audio cortex, along with the Convolutional Neural Networks (CNNs) of modern AI are very good at identifying the simple beauty or symmetry in these complex images or signals.

The beautiful simplicity of these textures is not always recognized.  My great grandmother was committed to an insane asylum in western Pennsylvania at 40 years of age where she remained for the rest of her life, because of the simple rambling textures of her speech, which the medical professionals found incoherent and the manifestation of insanity -- she was crazy.  She was from Transylvania and speaking in Romanian.  Are the hallucinations of schizophrenia, and the multiple personalities of Dissociative Identity Disorder just alternative textures of a high functioning creative mind, that can see the simple beauty of the maximalistic textures?  How often do we confuse genius with insanity?  Is insanity, really genius?  

Do hallucinogenic drugs give direct access to the high-level primal functionality of our brains, allowing us to better control our bodies and better see the beautiful simplicity of the world?  It has been established that people on hallucinogens perceive log-polar transforms \citep{ermentrout79}, like the HST shown in Eq.~\eqref{hst.eqn}.  The HST is a non-stationary, that is local, log-polar transform.  This can be seen by the fact that the activation function, $\ln(z)$, can be written as:  $\ln(\text{mod}(z)) + \text{i} \, \text{arg}(z)$, where $\ln(\text{mod}(z))$ is $\ln|z|$ or the log radius of $z$, and $\text{arg}(z)$ is the polar angle of $z$.  This is a foundational concept in computational neuroscience and mathematical psychology.

For complex or conservative collective systems there are two types of simply beautiful emergent behaviors.  The first comes from an inverse cascade of energy to larger scale and a relaxation of the system to self organized structures.  These are the local minimums, stable equilibriums, or o-points -- the simple geometric structures \citep{unser14,roy08} or minimalistic textures shown in Fig. \ref{simple.fig}.  The second comes from a normal cascade of energy to smaller scale and a relaxation of the system to a turbulent structure.  These are the local maximums, metastable equilibriums, or x-points -- the maximalistic, yet still simple, textures shown in Fig. \ref{textures.fig}.  Both are exposing the homology classes of the topology $\beta^*$.  The simple beauty is the symmetry or topology of the system quantified by the constants and their analytic continuation $H(\beta)$, the homology classes $\beta^*$, or the generators of the canonical flow and the constants $S_p[f(x)]$ and $H(p,q)$ or $S_P(q)$ and $E(P)$.  

These are all equivalent ways of quantifying the symmetry, that is the topology of the system.  Given the boundary $\beta^*$ where $\omega^*_Q=0$, that is the singular points or equilibriums, $H(\beta)$ can be found by solving the Laplace's equation.  Given the constants $H(p,q)=E(P)$, that is $\text{Re}(H(\beta))$, $\text{Im}(H(\beta))=Q=\partial S / \partial P$ can be found by analytic continuation.  Given the constants $H(p,q)$, the action $S_P(q)$ can be found by solving the HJB equation.  Remember that the action is also called the propagator, the generator of the canonical flow, the entropy, or the logarithm of the density.

\begin{figure}   
\noindent\includegraphics[width=\columnwidth]{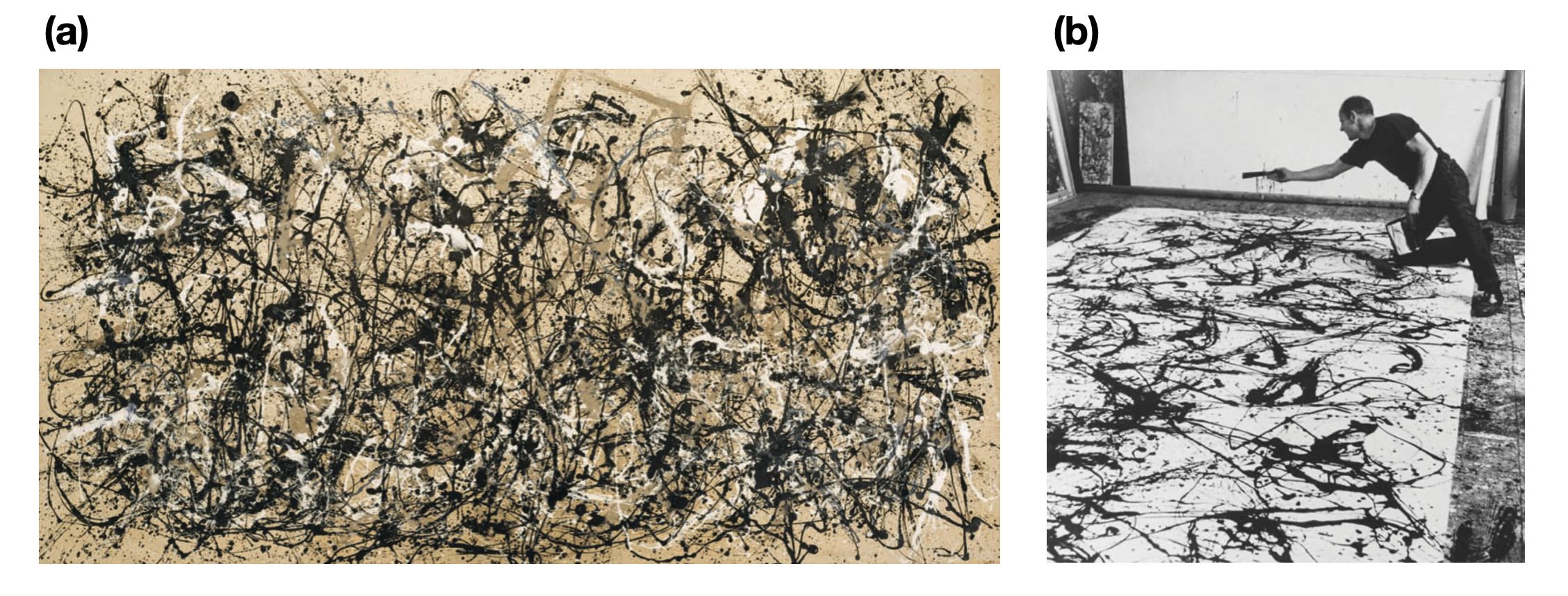}
\caption{\label{pollock.fig} Jackson Pollock action painting.  (a) An action painting by Jackson Pollock, \emph{Autumn Rhythm (Number 30)}. (b) Jackson Pollock action painting \emph{Autumn Rhythm}, photograph by Rudolph Burkhardt \href{https://youtu.be/zRE3J9fHpD8}{(YouTube video in Italian)}, and a film by Hans Namuth of Jackson Pollock painting \emph{Autumn Rhythm} \href{https://vimeo.com/428176462}{(Vimeo video of ``Autumn Rhythm in Summer'')}.}
\end{figure}

As a final observation, it is well known by art scholars \citep{stokstad.18} that Jackson Pollock's action paintings (see Fig. \ref{textures.fig}b and Fig. \ref{pollock.fig}) are not about the finished works, but are about the process or actions that were used to create them.  It is about the physical movements of the artist that threw, spattered, or dribbled the paint onto the canvas.  It is not about the final distribution or density of paint, it is about the process or flow or action that created it.  This is identical to the textures of collective systems.  It is not about the final distribution, but is about the generator of the canonical flow or action $S_P(q)$ that created it.  Ultimately, it is about the group symmetry of the painting or collective system that generates the canonical flow.

\begin{figure}   
\noindent\includegraphics[width=\columnwidth]{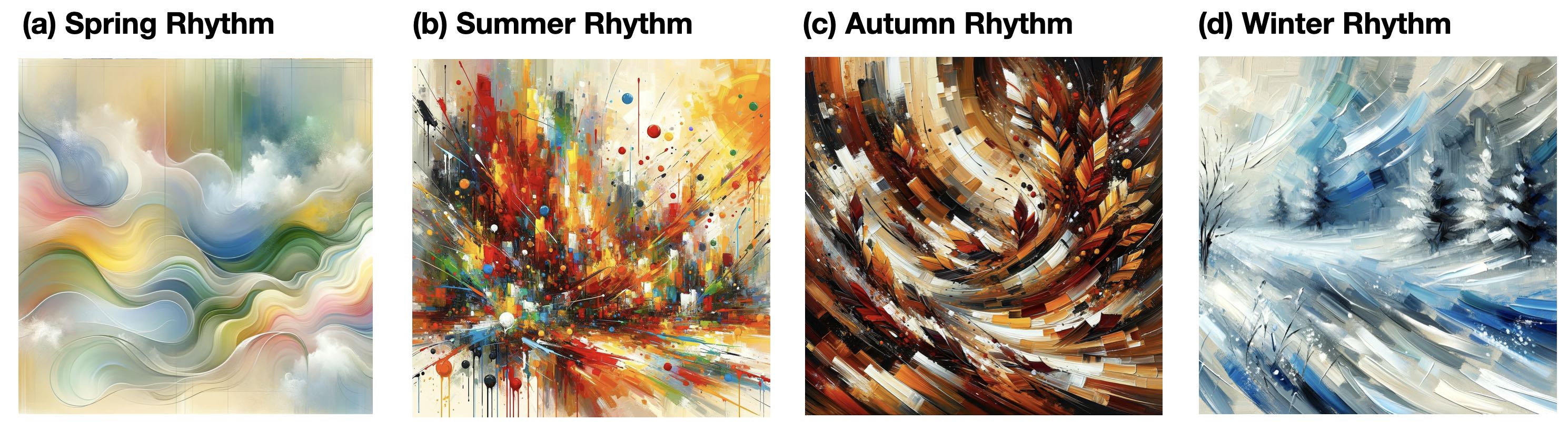}
\caption{\label{rhythms.fig} A series of abstract expressionistic action paintings generated by OpenAI's DALL-E, a GPT:  (a) \emph{Spring Rhythm}, (b) \emph{Summer Rhythm}, (c) \emph{Autumn Rhythm}, and (d) \emph{Winter Rhythm}.  Because of copyright restrictions, the paintings are generic ``completely abstract action paintings'', not ``in the style of Pollock''.}
\end{figure}

While it is easy to recreate a minimalistic painting like a Mondrian in the collective or field domain, a maximalistic painting like a Pollock is much easier to recreate in the process or dynamical domain of the puppet master or painter.  The deep deconvolution of the HST and the MLP decoder should be used to determine the motions or symmetries of the artist's movements, then the movements of the artist should be simulated and deeply convolved into the field domain of the painting.  While it is not necessary to do this process based recreation for a minimalistic painting, it can be used to generate a painting in the style of that artist that was not painted by that artist, but easily could have been painted by that artist.  This is a hallucination, not an exact copy.  This is what OpenAI's DALL-E has done in Fig. \ref{rhythms.fig}, creating the illustrations in an homage to the action painting of Pollock, a series of ``rhythms'' for each of the seasons.

Now we present an epilogue on the true Crazy Ones -- the ones that change things.  While well educated intelligence can learn the sticky textures $\beta^*$ from historical observation, it can not create new sticky textures.  It can not create an abstract action painting if none has yet been painted.  It takes theoretical play where new sticky textures are proposed, then the ramifications of those additional sticky textures explored through play.  This is how the true creative Crazy Ones innovate and change things.  They ``think different'', the title of the famous 1997 Apple ad campaign that talked about the Crazy Ones who change things and move the human race forward.  The creative art or true genius is the choice of the new sticky textures, and the creative process is the theoretical play.  So, although we are at the dawn of well educated artificial intelligence, we still must rely upon organic human creativity -- the Crazy Ones!

\end{document}